\documentclass[11pt]{article}

\usepackage[preprint]{acl}

\usepackage{times}
\usepackage{latexsym}

\usepackage[T1]{fontenc}

\usepackage[utf8]{inputenc}

\usepackage{microtype}

\usepackage{inconsolata}

\usepackage{graphicx}

\usepackage{amsmath}
\usepackage{amssymb}
\usepackage{mathtools}
\usepackage{amsthm}
\usepackage{booktabs}
\usepackage{multirow}
\usepackage{colortbl}
\usepackage{xcolor}
\usepackage{enumitem}
\definecolor{graybg}{gray}{0.95} 

\usepackage[capitalize,noabbrev]{cleveref}

\theoremstyle{plain}

\usepackage[most]{tcolorbox}
\usepackage{listings}
\tcbuselibrary{listings,breakable}

\newtcblisting{QEMempromptbox}[1]{
    enhanced,
    breakable,
    listing only,
    listing engine=listings,
    colback=black!2,
    colframe=black!40,
    boxrule=0.5pt,
    arc=2pt,
    left=2mm,
    right=2mm,
    top=1mm,
    bottom=1mm,
    fonttitle=\bfseries\small,
    title={#1},
    listing options={
        basicstyle=\ttfamily\footnotesize,
        breaklines=true,
        breakatwhitespace=true,
        columns=fullflexible,
        keepspaces=true,
        showstringspaces=false
    }
}

\title{From Relevant Memories to Sufficient Evidence: Rethinking Memory Utilization in LLM Agents}

\author{
 \textbf{Shixuan Li\textsuperscript{1}\thanks{These authors contributed equally to this work.}},
 \textbf{Wei Yang\textsuperscript{1}\footnotemark[1]},
 \textbf{Peiyu Zhang\textsuperscript{1}},
 \textbf{Anzhe Cheng\textsuperscript{1}},
 \textbf{Heng Ping\textsuperscript{1}},
\\
 \textbf{Xiaole Zhang\textsuperscript{1}},
 \textbf{Jesse Thomason\textsuperscript{2}},
 \textbf{Xuezhe Ma \textsuperscript{1}},
 \textbf{Yue Zhao\textsuperscript{1}},
 \textbf{Paul Bogdan\textsuperscript{1}},
\\
 \textsuperscript{1}University of Southern California
 \textsuperscript{2}Georgia Institute of Technology
}

\begin{document}
\maketitle
\begin{abstract}
Long-term memory enables LLM agents to reuse past experiences, yet retrieving relevant memories does not necessarily provide sufficient evidence for answering a query. Essential information may remain fragmented across memory records, obscured by abstraction, or disconnected from the relations required for reasoning.
We identify this \emph{relevance--sufficiency gap} and propose QEMem, a framework that reformulates memory utilization as the construction of query-specific evidence sets, shifting the focus from individual retrieval relevance to collective evidential support.
QEMem integrates three complementary mechanisms: query-guided discovery retrieves evidence across multiple memory granularities; source-grounded reification organizes it into structured ViewCards with explicit evidential roles and source references; and sufficiency-guided refinement selectively recovers missing information.
Operating on existing memory stores without modifying memory formation, QEMem achieves 91.43\% LLM-judge accuracy on LoCoMo with Claude Opus 4.5, outperforming the strongest evaluated baseline, MemR$^3$, by 7.6 percentage points and NEMORI by 10.7 points under a shared memory store.
Further analyses demonstrate improved evidence coverage and strong performance on questions requiring cross-session evidence integration. Code is available at \url{https://github.com/shixuanli01/QEMem}.
\end{abstract}

\section{Introduction}

Long-term memory enables large language model (LLM) agents to maintain continuity across extended interactions and reuse past experiences. Recent systems, including Mem0~\citep{chhikara2025mem0}, A-Mem~\citep{xu2025mem}, and NEMORI~\citep{nan2025nemori}, have advanced the persistent storage and structured organization of memories, while SimpleMem~\citep{liu2026simplemem} and related approaches improve memory consolidation and retrieval. Although these advances make past information increasingly accessible, retrieving relevant memories does not necessarily provide sufficient evidence to answer a query.

A central difficulty lies in the mismatch between the granularity of retrieved memories and the facts and relations required to answer a query. Relevant information may remain fragmented across memory records or obscured by abstraction, while the relationships needed to connect individual facts may remain implicit. Recent approaches, including MemGAS~\citep{xu2026single}, REMem~\citep{shu2026remem}, RaMem~\citep{yang2026ramem} and MemR$^3$~\citep{du2025memr}, have advanced multi-granular memory access, episodic reasoning, and reflective retrieval, respectively. Even with these advances, individually relevant memories may still fail to jointly provide the facts and relations needed for answering. We characterize this disconnect as the \emph{relevance--sufficiency gap}, raising a fundamental question:
How can an agent construct a source-grounded evidence set that collectively satisfies a query's information requirements?

To address this question, we shift the fundamental unit of memory utilization from individual memory records to a \emph{query-specific evidence set}, whose value lies in the joint support of complementary evidence rather than isolated retrieval relevance. Constructing such a set requires recovering the necessary facts and relations, preserving their source references, and identifying missing information within limited retrieval and reasoning budgets. Our diagnostic analysis illustrates why retaining past information is not enough to make the required evidence accessible. On LoCoMo~\citep{maharana2024evaluating}, although NEMORI retains nearly all original interaction records, its retrieved episodes cover the source turns of all annotated supporting evidence for only 68.8\% of questions. On questions with complete source coverage, answer accuracy reaches approximately 93\% (Section~\ref{sec:rq2}), suggesting that access to stored evidence remains a major bottleneck even for strong LLMs. These findings motivate \emph{query-specific evidence construction} as a central objective of long-term memory utilization.

To operationalize this perspective, we propose QEMem, a framework that constructs query-specific evidence sets over an existing memory store. QEMem begins with \emph{query-guided evidence discovery}, identifying the information requirements of a query and recovering complementary evidence across episodic memories, semantic facts, and original interaction turns. Through source-grounded evidence reification, retrieved fragments are organized into structured ViewCards that make query-relevant claims, evidential relations, temporal information, and supporting source spans explicit. Finally, sufficiency-guided evidence refinement identifies unresolved information needs and selectively retrieves additional evidence, providing a more complete basis for answer generation.

Our main contributions are:
\begin{itemize}
    \item[(i)] We identify the \emph{relevance--sufficiency gap} in long-term agent memory and introduce the \emph{query-specific evidence set} as a perspective that shifts memory utilization from individual relevance to collective evidential support.

    \item[(ii)] We propose \textbf{QEMem}, a framework that unifies query-guided evidence discovery, source-grounded evidence reification, and sufficiency-guided refinement to construct structured, source-grounded evidence for reasoning.

    \item[(iii)] Comprehensive experiments on LoCoMo demonstrate that QEMem achieves 91.43\% LLM-judge accuracy, outperforming the strongest evaluated baseline, MemR$^3$, by 7.6 percentage points and NEMORI by 10.7 points under an identical memory store.
\end{itemize}

\section{Related Work}
\label{sec:related_work}

Our work connects long-term agent memory~\citep{park2023generativeagentsinteractivesimulacra,hu2025memory} with retrieval and evidence reasoning, focusing on how stored information is assembled into query-specific evidence.

\textbf{Long-Term Memory for LLM Agents.}
MemGPT~\citep{packer2024memgptllmsoperatingsystems}, MemoryBank~\citep{zhong2024memorybank}, and MemoryOS~\citep{kang2025memory} support persistent storage and hierarchical memory management.
Mem0~\citep{chhikara2025mem0} extracts reusable facts, while A-Mem~\citep{xu2025mem} and Zep~\citep{rasmussen2025zep} organize interconnected knowledge.
LightMem~\citep{fang2025lightmem}, SimpleMem~\citep{liu2026simplemem}, and MemGAS~\citep{xu2026single} improve consolidation and multi-granular access.
NEMORI~\citep{nan2025nemori}, AgeMem~\citep{yu2026agentic}, and MemAgent~\citep{yu2025memagent} further explore adaptive retention and memory operations.
These advances improve memory storage and access, while QEMem focuses on assembling complementary evidence for a given query.

\textbf{Retrieval and Evidence Reasoning.}
Retrieval-augmented generation~\citep{lewis2020retrieval} accesses external knowledge through dense and lexical retrieval~\citep{karpukhin2020densepassageretrievalopendomain,10.1561/1500000019}, although additional context does not necessarily improve evidence utilization~\citep{liu2024lost}.
GraphRAG~\citep{edge2024local} and LightRAG~\citep{guo2024lightrag} exploit structured knowledge, while ReAct~\citep{yao2022react}, Self-RAG~\citep{asai2023self}, and Search-R1~\citep{jin2025searchr1trainingllmsreason} enable adaptive information acquisition.
For conversational memory, MemR$^3$~\citep{du2025memr} combines retrieval and reflection, REMem~\citep{shu2026remem} supports graph-guided episodic reasoning, and MRAgent~\citep{ji2026memory} and E-mem~\citep{wang2026mem} explore memory reconstruction.
As conversational evidence spans sessions, temporal contexts, and abstraction levels~\citep{maharana2024evaluating,jia-etal-2025-evaluating}, QEMem explicitly organizes retrieved fragments into source-grounded evidence sets and refines them to address missing information.
Further discussion appears in Appendix~\ref{app:related_work}.

\section{Problem Formulation}
\label{sec:problem_formulation}

Given a long-term interaction history $\mathcal{D}=\{u_1,\ldots,u_T\}$, a memory backbone constructs episodic memories $\mathcal{M}_{E}$ and semantic memories $\mathcal{M}_{S}$, while retaining original turns $\mathcal{M}_{U}$ through source references.
The resulting memory space is $\mathcal{M}=\{\mathcal{M}_{E},\mathcal{M}_{S},\mathcal{M}_{U}\}$.
For a query $q$, individually relevant records may provide only part of the evidence needed for answering.

We therefore formulate memory utilization as constructing a \emph{query-specific evidence set} $\mathcal{E}_q$ that collectively supports the information requirements of $q$.
Let $\mathcal{C}(q,\mathcal{M})$ denote candidate evidence derivable from the stored memories.
Our conceptual objective is
\begin{equation}
\label{eq:evidence_objective}
\begin{aligned}
\mathcal{E}_q^{*}
&= \arg\max_{\mathcal{E}\subseteq\mathcal{C}(q,\mathcal{M})}
\operatorname{Suff}(q,\mathcal{E})\\
&\text{s.t.}\quad
\operatorname{Traceable}(\mathcal{E},\mathcal{M}),\quad
\operatorname{Cost}(\mathcal{E})\leq B,
\end{aligned}
\end{equation}
where $\operatorname{Suff}$ measures collective evidential support,
$\operatorname{Traceable}$ requires identifiable source references,
and $B$ bounds retrieval and reasoning cost.

\section{Method}
\label{sec:method}

\begin{figure*}[t]
    \centering
    \includegraphics[width=\textwidth]{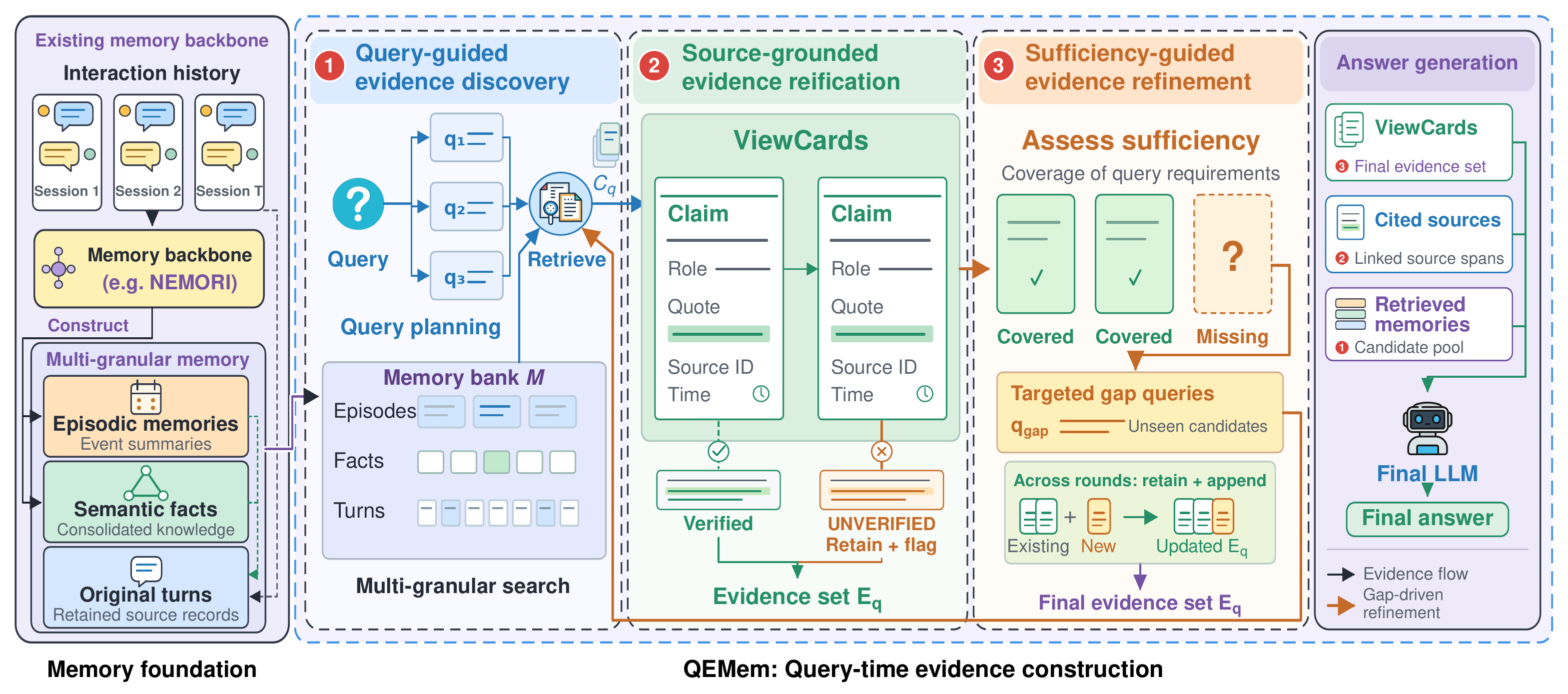}
    \caption{
    \textbf{Overview of QEMem.}
    The framework addresses the relevance--sufficiency gap by moving from individually relevant memories to collectively sufficient evidence.
    Given a query, QEMem discovers candidate evidence across multiple memory granularities, constructs source-grounded evidence representations, and iteratively refines the evidence set based on identified information gaps.
    The resulting query-specific evidence set supports grounded answer generation.
    }
    \label{fig:method_overview}
\end{figure*}

We propose \textbf{QEMem}, a framework that reformulates long-term memory utilization as query-conditioned evidence-set construction, as illustrated in Figure~\ref{fig:method_overview}. Rather than directly optimizing the conceptual objective in Eq.~\eqref{eq:evidence_objective}, QEMem progressively constructs an evidence set through three complementary mechanisms: \emph{query-guided evidence discovery}, \emph{source-grounded evidence reification}, and \emph{sufficiency-guided evidence refinement}. As illustrated in Figure~\ref{fig:method_overview}, these mechanisms transform heterogeneous memory representations into complementary, source-grounded evidence that collectively supports the query, without modifying the underlying memory formation process. 

\subsection{Query-Guided Evidence Discovery}
\label{sec:evidence_discovery}

Constructing a sufficient evidence set requires recovering complementary information that may be distributed across memory representations and interaction sessions. Conventional relevance-based retrieval ranks individual memories independently, potentially overlooking evidence whose contribution becomes apparent only when combined with other facts. QEMem therefore conditions evidence discovery on the information requirements of the query rather than its surface-level semantic similarity alone.

\paragraph{Query-Conditioned Retrieval Planning.}
Given a query $q$, a planner $\mathcal{P}$ induces a structured information requirement:
\begin{equation}
\label{eq:query_planning}
\mathcal{P}(q)
=
\left(
\eta_q, a_q, \alpha_q, \tau_q, \mathcal{Q}_q
\right),
\end{equation}
where $\eta_q$ specifies the information need, $a_q$ denotes the expected answer type, and $\alpha_q$ and $\tau_q$ indicate requirements for cross-memory aggregation and temporal reasoning, respectively. The query set $\mathcal{Q}_q$ comprises the original query and complementary retrieval queries targeting semantic reformulations and intermediate information needs. This decomposition broadens evidence discovery beyond memories directly matching the original query.

\paragraph{Multi-Granular Evidence Retrieval.}
QEMem searches three complementary memory spaces $\mathcal{M}_g$, where $g\in\mathcal{G}=\{E,S,U\}$ denotes episodic memories, semantic facts, and original interaction turns, respectively. For each $q_j\in\mathcal{Q}_q$, dense and sparse retrieval produce ranked candidate lists $\mathcal{L}_{j,g}^{h}$, with $h\in\{\mathrm{dense},\mathrm{sparse}\}$. We aggregate these retrieval signals through reciprocal rank fusion:
\begin{equation}
\label{eq:rrf}
R_g(m\mid q)
=
\sum_{q_j\in\mathcal{Q}_q}
\sum_h
\frac{
\mathbb{I}[m\in\mathcal{L}_{j,g}^{h}]
}{
k+\operatorname{rank}_{\mathcal{L}_{j,g}^{h}}(m)
},
\end{equation}
where $k$ controls the contribution of individual rankings and absent candidates contribute zero.

To account for dependencies across memory granularities, provenance links between semantic facts and their originating episodes further inform candidate ranking. Let $\widetilde{R}_g$ denote the resulting ranking score, with provenance adjustment applied to episodic memories. The initial candidate pool is constructed as
\begin{equation}
\label{eq:evidence_discovery}
\mathcal{C}_q^{(0)}
=
\operatorname{Ctx}_{U}
\left(
\bigcup_{g\in\mathcal{G}}
\operatorname{Top}_{b_g(q)}
\left(\mathcal{M}_g;\widetilde{R}_g\right)
\right),
\end{equation}
where $b_g(q)$ specifies the granularity-dependent retrieval budget and $\operatorname{Ctx}_{U}$ augments retrieved interaction turns with local conversational context. Source-level retrieval budgets adapt to aggregation requirements, allowing broader coverage when relevant facts may span multiple events or sessions. The resulting $\mathcal{C}_q^{(0)}$ preserves both abstract memory representations and fine-grained source observations for subsequent evidence construction.

\subsection{Source-Grounded Evidence Reification}
\label{sec:evidence_reification}

Evidence discovery identifies potentially useful memory fragments, but their relationships to the query remain implicit. QEMem addresses this limitation through \emph{source-grounded evidence reification}, transforming retrieved information into structured evidence objects whose roles and supporting sources are explicitly represented.

\paragraph{Query-Conditioned Evidence Construction.}
Given $\mathcal{C}_q^{(0)}$ and the information requirement $\eta_q$, an evidence constructor $\mathcal{R}$ jointly produces an initial evidence set and an evidence-gap assessment:
\begin{equation}
\label{eq:evidence_reification}
\left(
\mathcal{E}_q^{(0)},\mathbf{f}_q^{(0)}
\right)
=
\mathcal{R}
\left(
q,\eta_q,\mathcal{C}_q^{(0)};\varnothing
\right),
\end{equation}
where $\mathcal{E}_q^{(0)}=\{v_i\}_{i=1}^{N_q}$ contains the constructed evidence objects, and $\mathbf{f}_q^{(0)}$ provides feedback for subsequent refinement.

Each evidence object is represented as a structured \emph{ViewCard}:
\begin{equation}
\label{eq:viewcard}
v_i =
\left(
\gamma_i,r_i,\pi_i,\xi_i,t_i
\right),
\end{equation}
where $\gamma_i$ denotes a query-relevant claim, $r_i$ specifies its evidential relation, $\pi_i$ identifies supporting memory sources, $\xi_i$ preserves a verbatim source span, and $t_i$ represents temporal information when applicable.

The relation $r_i$ characterizes how the evidence contributes to the query, including answer candidates, intermediate facts, temporal constraints, and inference support. Unlike ordinary memory summarization, this representation explicitly connects individual observations to their query-specific evidential functions, allowing complementary information from multiple memories to be organized within a common evidence set.

\paragraph{Source Grounding and Temporal Alignment.}
To preserve traceability, QEMem assesses the correspondence between each ViewCard's quotation and its cited sources:
\begin{equation}
\label{eq:source_verification}
\begin{aligned}
\kappa_i
&=
\max_{m\in\operatorname{Src}(\pi_i)}
\operatorname{Match}(\xi_i,x_m),\\
\nu_i
&=
\mathbb{I}[\kappa_i\geq\delta],
\end{aligned}
\end{equation}
where $x_m$ denotes the source text, $\operatorname{Match}$ measures textual correspondence, and $\nu_i$ indicates whether the quoted evidence is source-verified. This procedure establishes quotation fidelity rather than logical entailment of the structured claim.

For temporally grounded evidence, relative expressions are further interpreted against their originating interaction timestamps, with deterministic normalization applied when appropriate. Unverified quotations are flagged rather than discarded, while conflicting observations remain available for downstream reasoning. Consequently, $\mathcal{E}_q^{(0)}$ provides a structured, traceable representation in which retrieved information is organized by its evidential contribution to the query.

\subsection{Sufficiency-Guided Evidence Refinement}
\label{sec:evidence_refinement}

Although evidence reification makes supporting relations explicit, the assembled evidence may still omit information required for answering. QEMem therefore performs \emph{sufficiency-guided evidence refinement}, using the current evidence set to identify unresolved information needs and guide targeted retrieval.

\paragraph{Evidence Sufficiency Assessment.}
At refinement step $t$, the evidence constructor maintains feedback
\begin{equation}
\label{eq:sufficiency_assessment}
\mathbf{f}_q^{(t)}
=
\left(
\hat{s}_q^{(t)},
d_q^{(t)},
\mathcal{Q}_{q,\mathrm{gap}}^{(t)}
\right),
\end{equation}
where $\hat{s}_q^{(t)}\in\{0,1\}$ indicates estimated evidential sufficiency, $d_q^{(t)}$ describes unresolved information requirements, and $\mathcal{Q}_{q,\mathrm{gap}}^{(t)}$ contains targeted retrieval queries addressing those gaps. The assessment considers whether the evidence set collectively supports the query, including the intermediate facts and cross-memory relations necessary for multi-hop and aggregate reasoning. This feedback is generated jointly with the ViewCards rather than by a separately trained sufficiency model.

\paragraph{Evidence-Gap-Conditioned Refinement.}
When $\hat{s}_q^{(t)}=0$, QEMem retrieves additional evidence according to the identified gaps, excluding previously examined candidates:
\begin{equation}
\label{eq:refinement_retrieval}
\Delta\mathcal{C}_q^{(t+1)}
=
\operatorname{Retrieve}
\left(
\mathcal{Q}_{q,\mathrm{gap}}^{(t)},
\mathcal{M}
\right)
\setminus
\mathcal{C}_q^{(t)}.
\end{equation}
Rather than reconstructing the entire evidence set, QEMem conditions subsequent evidence construction on previously established ViewCards:
\begin{equation}
\label{eq:evidence_update}
\begin{aligned}
\left(
\Delta\mathcal{E}_q^{(t+1)},
\mathbf{f}_q^{(t+1)}
\right)
&=
\mathcal{R}
\left(
q,\eta_q,\Delta\mathcal{C}_q^{(t+1)};
\mathcal{E}_q^{(t)}
\right),\\
\mathcal{E}_q^{(t+1)}
&=
\mathcal{E}_q^{(t)}
\cup
\Delta\mathcal{E}_q^{(t+1)}.
\end{aligned}
\end{equation}
Here, $\Delta\mathcal{E}_q^{(t+1)}$ contains newly constructed evidence, while the existing set is preserved. The retrieved candidate pool is updated analogously. This refinement process progressively incorporates complementary evidence without repeatedly processing previously examined memories. Iteration terminates when sufficiency is estimated to be satisfied, no new candidates are available, or the retrieval budget is exhausted.

\paragraph{Evidence-Grounded Answer Generation.}
Let $\mathcal{E}_q^{(T)}$ denote the evidence set obtained after refinement. The final answer is generated by a frozen LLM $F_\theta$:
\begin{equation}
\label{eq:grounded_generation}
\hat{y}
=
F_\theta
\left(
q,
\operatorname{Assemble}
\left(
\mathcal{E}_q^{(T)},
\mathcal{C}_q^{(T)}
\right)
\right),
\end{equation}
where $\operatorname{Assemble}$ presents the ViewCards together with their supporting source context and remaining retrieved memories. This evidence-first organization exposes the collective support of the assembled evidence while preserving access to original observations for verification and disambiguation. Sufficiency assessments remain approximate, serving as feedback for evidence refinement rather than guaranteeing formal completeness.

\section{Experiments}

\textbf{Experimental setup.}
We compare QEMem with seven baselines on 1,540 LoCoMo questions~\citep{maharana2024evaluating}, using Haiku 4.5 and Opus 4.5. Details appear in Appendix~\ref{app:experimental_setup}.

\subsection{RQ1: Overall Performance}
\label{sec:rq1}

QEMem achieves the highest overall accuracy (Table~\ref{tab:rq1_main}): \textbf{89.61\%} with Haiku 4.5 and \textbf{91.43\%} with Opus 4.5, surpassing the strongest baseline, MemR$^3$, by 9.16 and 7.60 percentage points, respectively.
Both gains are significant under paired evaluation ($p<10^{-5}$; 95\% CIs: [7.1, 11.2] and [5.8, 9.3] points).
Gains are pronounced on multi-hop questions (+8.16 and +10.29 points over SimpleMem) and temporal reasoning (+6.23 and +4.36 points over MemR$^3$).
These improvements are consistent with more effective recovery and integration of complementary evidence across memory records.
QEMem leads all four categories under Haiku and three under Opus, where open-domain reasoning is the only exception.
\textbf{Performance across backbones.}
Switching from Opus to Haiku reduces QEMem's accuracy by only 1.82 points, compared with 4.22 for NEMORI and 3.38 for MemR$^3$.
QEMem with Haiku outperforms every baseline using Opus, supporting its effectiveness across both backbones.
Its gains over NEMORI under an identical memory store highlight the value of improving query-time memory utilization.



\begin{table*}[t]
\centering
\caption{
Main results on LoCoMo across two Claude backbones, measured by LLM-judge accuracy (\%).
All memory stores are constructed using Claude Opus 4.5, while query-time processing uses the indicated backbone.
\textbf{Bold} and \underline{underlined} values indicate the best and second-best results within each backbone, respectively.
}
\label{tab:rq1_main}
\setlength{\tabcolsep}{2.36pt}
\renewcommand{\arraystretch}{1.0}
\small

\begin{tabular*}{\textwidth}{@{\extracolsep{\fill}}lccccc@{\hspace{10pt}}ccccc@{}}
\toprule
& \multicolumn{5}{c}{\textbf{Claude Haiku 4.5}}
& \multicolumn{5}{c}{\textbf{Claude Opus 4.5}} \\
\cmidrule(lr){2-6}
\cmidrule(lr){7-11}
\textbf{Method}
& \textbf{Multi-Hop}
& \textbf{Temporal}
& \textbf{Open}
& \textbf{Single-Hop}
& \textbf{Overall}
& \textbf{Multi-Hop}
& \textbf{Temporal}
& \textbf{Open}
& \textbf{Single-Hop}
& \textbf{Overall} \\
\midrule

NEMORI
& 69.50 & 71.65 & 50.00 & 83.71 & 76.49
& 70.92 & 77.88 & 61.46 & 87.28 & 80.71 \\

Mem0
& 65.25 & 61.68 & 52.08 & 74.67 & 68.83
& 62.77 & 68.85 & 62.50 & 80.26 & 73.57 \\

A-Mem
& 50.35 & 32.09 & 44.79 & 68.73 & 56.23
& 51.77 & 29.28 & 48.96 & 73.60 & 58.83 \\

LightMem
& 63.48 & 81.62 & 57.29 & 78.24 & 74.94
& 62.41 & 84.42 & 59.38 & 80.14 & 76.49 \\

SimpleMem
& \underline{78.72} & 79.13 & 53.12 & 80.62 & 78.25
& \underline{78.01} & 81.62 & 59.38 & 84.54 & 81.17 \\

MemR$^3$
& 58.87 & \underline{82.87} & \underline{61.46}
& \underline{88.94} & \underline{80.45}
& 62.41 & \underline{86.92} & \textbf{76.04}
& \underline{90.73} & \underline{83.83} \\

MemGAS
& 36.88 & 52.34 & 47.92 & 73.37 & 60.71
& 41.13 & 56.70 & 52.08 & 72.53 & 62.21 \\

\midrule
\textbf{QEMem (Ours)}
& \textbf{86.88}
& \textbf{89.10}
& \textbf{68.75}
& \textbf{93.10}
& \textbf{89.61}
& \textbf{88.30}
& \textbf{91.28}
& \underline{69.79}
& \textbf{95.01}
& \textbf{91.43} \\

\bottomrule
\end{tabular*}
\end{table*}

\subsection{RQ2: Evidence Coverage and Answer Accuracy}
\label{sec:rq2}

\begin{figure*}[t]
    \centering
    \includegraphics[width=\textwidth]{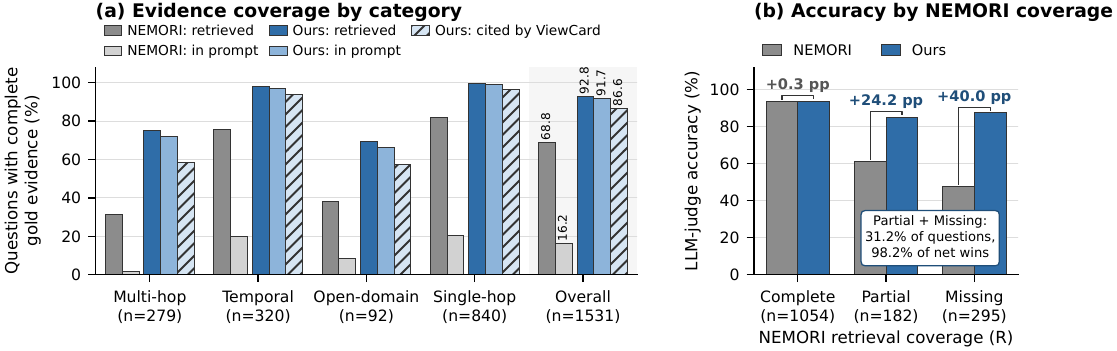}
\caption{
    \textbf{Evidence coverage and answer accuracy on LoCoMo.}
    \textbf{(a)} Complete coverage of annotated supporting turns.
    Retrieval-source coverage ($R$) includes turns belonging to retrieved episodes or present in the final raw-turn context.
    Source visibility ($V$) counts original turns included verbatim in the answer prompt.
    Card-source coverage ($C$) includes directly cited turns and all turns belonging to cited episodes.
    Coverage is complete when all annotated supporting turns are covered.
    \textbf{(b)} Answer accuracy grouped by NEMORI's retrieval-source coverage.
    Results use Opus 4.5 on 1,531 questions with fully mappable evidence annotations.
    }
    \label{fig:rq2_evidence}
\end{figure*}

Figure~\ref{fig:rq2_evidence} examines how annotated evidence coverage relates to answer accuracy.
\textbf{Evidence coverage.}
With Opus 4.5, QEMem increases complete retrieval-source coverage from 68.8\% to 92.8\% and complete source visibility from 16.2\% to 91.7\% (Figure~\ref{fig:rq2_evidence}(a)).
ViewCard sources cover all annotated supporting turns for 86.6\% of questions through direct turn citations or provenance links through cited episodes.
On multi-hop questions, complete source visibility rises from 1.8\% to 72.0\%, improving exposure of distributed evidence.
\textbf{Answer accuracy.}
Figure~\ref{fig:rq2_evidence}(b) groups questions by NEMORI's retrieval-source coverage.
The methods achieve comparable accuracy in the complete-coverage group, while QEMem improves accuracy by 24.2 and 40.0 percentage points in the partial- and missing-coverage groups, respectively.
These groups comprise 31.2\% of questions but account for 98.2\% of the net accuracy gain, showing that QEMem's improvements concentrate on questions with incomplete baseline evidence coverage.
This pattern supports evidence recovery as a major bottleneck in long-term memory reasoning.
Haiku 4.5 shows a similar pattern (Appendix~\ref{app:rq2_haiku}).

\subsection{RQ3: Dissecting Query-Specific Evidence Construction}
\label{sec:rq3}

Table~\ref{tab:rq3_ablation} compares the stages of query-specific evidence construction.
Under matched final answer-context budgets, Full QEMem outperforms retrieval-only answering by 5.9 and 5.7 percentage points with Haiku and Opus, respectively ($p<10^{-10}$ for both).
\textbf{Evidence discovery.}
Query-guided multi-granular retrieval with direct answering improves accuracy over matched-budget retrieval by 4.7 points with Haiku and 5.2 points with Opus, accounting for most of the overall gain.
\textbf{Evidence reification.}
The reification stage constructs and verifies ViewCards and expands their cited source context.
Compared with direct answering, this stage increases complete source visibility from 82.6\% to 89.3\% with Haiku and from 83.1\% to 90.7\% with Opus.
Accuracy changes by $+0.7$ and $-0.3$ points, respectively.
Its clearest benefit in this comparison is greater exposure of annotated source evidence through card-guided context expansion.
\textbf{Evidence refinement.}
Refinement further improves overall accuracy by 0.5 points with Haiku and 0.8 points with Opus.
On the same 131 Opus questions that trigger refinement, accuracy rises from 61.8\% to 71.0\% ($+9.2$ points, paired McNemar $p=0.0075$).
The corresponding gain for Haiku is positive but not statistically significant.
Appendix~\ref{app:rq5_efficiency} provides reconstruction details, sensitivity analysis, and accuracy--cost comparisons.

\begin{table*}[t]
\centering
\caption{
\textbf{Ablation analysis of query-specific evidence construction on LoCoMo.}
Accuracy (\%) is evaluated on 1,540 questions;
$\Delta$ is the difference from Full QEMem in percentage points.
Visibility is the percentage of questions with all annotated supporting turns exposed verbatim in the final answer context, evaluated on 1,531 questions with fully mappable annotations.
Matched-Budget Retrieval matches Full QEMem's final answer-context token budget.
Bold indicates the best result in each accuracy and visibility column.
}
\label{tab:rq3_ablation}
\setlength{\tabcolsep}{5pt}
\renewcommand{\arraystretch}{1.0}
\footnotesize

\textbf{(a) Claude Haiku 4.5}
\vspace{3pt}
\begin{tabular*}{\textwidth}{@{\extracolsep{\fill}}lccccccc@{}}
\toprule
\textbf{Variant}
& \textbf{Multi-Hop}
& \textbf{Temporal}
& \textbf{Open-Domain}
& \textbf{Single-Hop}
& \textbf{Overall}
& $\boldsymbol{\Delta}$
& \textbf{Visibility} \\
\midrule
Matched-Budget Retrieval
& 79.1 & 80.4 & 64.6 & 88.7 & 83.7 & $-5.9$ & 77.0 \\
Query-Guided Retrieval
& 84.4 & 85.0 & 66.7 & 93.5 & 88.4 & $-1.2$ & 82.6 \\
w/o Evidence Refinement
& 84.0 & 88.2 & 67.7 & \textbf{93.6}
& 89.1 & $-0.5$ & 89.3 \\
\midrule
\textbf{Full QEMem}
& \textbf{86.9} & \textbf{89.1} & \textbf{68.8}
& 93.1 & \textbf{89.6} & -- & \textbf{91.6} \\
\bottomrule
\end{tabular*}
\vspace{5pt}
\textbf{(b) Claude Opus 4.5}
\vspace{3pt}
\begin{tabular*}{\textwidth}{@{\extracolsep{\fill}}lccccccc@{}}
\toprule
\textbf{Variant}
& \textbf{Multi-Hop}
& \textbf{Temporal}
& \textbf{Open-Domain}
& \textbf{Single-Hop}
& \textbf{Overall}
& $\boldsymbol{\Delta}$
& \textbf{Visibility} \\
\midrule
Matched-Budget Retrieval
& 80.1 & 83.8 & \textbf{72.9} & 89.8
& 85.7 & $-5.7$ & 75.0 \\
Query-Guided Retrieval
& 86.9 & 89.4 & 71.9 & \textbf{95.0}
& 90.9 & $-0.5$ & 83.1 \\
w/o Evidence Refinement
& 86.9 & 90.0 & 69.8 & 94.5
& 90.6 & $-0.8$ & 90.7 \\
\midrule
\textbf{Full QEMem}
& \textbf{88.3} & \textbf{91.3} & 69.8
& \textbf{95.0} & \textbf{91.4} & --
& \textbf{91.7} \\
\bottomrule
\end{tabular*}
\end{table*}

\subsection{RQ4: Source Grounding and Evidential Reliability}
\label{sec:rq4}

We evaluate the source traceability and quotation fidelity of ViewCards, together with the groundedness of generated outputs.
\textbf{ViewCard grounding.}
Table~\ref{tab:rq4_grounding} reports results over all 10,121 Haiku and 10,213 Opus ViewCards.
Source references resolve to valid memory records for 98.4\% and 100.0\% of cards, respectively.
Normalized quotation fidelity reaches 85.5\% and 97.7\%, increasing to 90.0\% and 99.1\% with segment-wise verification of multi-excerpt quotations.
All cards citing original dialogue turns are traceable to their source identifiers.
These results show that ViewCards make evidence provenance and quotation fidelity directly auditable.
Additional claim--source consistency and temporal-grounding analyses are provided in Appendix~\ref{app:rq4_grounding}.
\textbf{Output grounding.}
On 240 stratified questions evaluated with Opus, QEMem improves full-output groundedness over NEMORI and MemR$^3$ by 15.5 and 11.3 percentage points, respectively, while maintaining a lower abstention rate.
This audit evaluates complete model outputs, including auxiliary reasoning and explanations.
Appendix~\ref{app:answer_groundedness} reports detailed comparisons, core-answer grounding, and the relationship between output length and groundedness.

\begin{table}[t]
\centering
\caption{
\textbf{Source traceability and quotation fidelity of QEMem ViewCards on LoCoMo.}
Percentages are computed over all generated cards
(10,121 for Haiku and 10,213 for Opus),
except original-turn traceability, which uses cards citing original dialogue turns.
Multi-excerpt fidelity permits segment-wise quotation matching.
}
\label{tab:rq4_grounding}
\setlength{\tabcolsep}{4pt}
\renewcommand{\arraystretch}{1.1}
\footnotesize
\begin{tabular}{lcc}
\toprule
\textbf{Grounding Metric}
& \textbf{Haiku 4.5}
& \textbf{Opus 4.5} \\
\midrule
Provenance Resolution
& 98.4 & \textbf{100.0} \\
Exact Quotation Fidelity
& 82.7 & \textbf{95.9} \\
Normalized Quotation Fidelity
& 85.5 & \textbf{97.7} \\
Multi-Excerpt Quotation Fidelity
& 90.0 & \textbf{99.1} \\
Original-Turn Traceability
& \textbf{100.0} & \textbf{100.0} \\
\bottomrule
\end{tabular}
\end{table}

\subsection{RQ5: Reasoning over Complex and Distributed Evidence}
\label{sec:rq5}

Table~\ref{tab:rq5_complexity} evaluates questions requiring multiple supporting turns, cross-session evidence, and cross-session list/count aggregation.
\textbf{QEMem achieves the highest accuracy in every reported group under both backbones.}
On questions with at least three supporting turns, it reaches 81.0\% accuracy with Haiku and 82.1\% with Opus, exceeding the strongest included baseline by 6.0 and 8.2 percentage points, respectively.
\textbf{The gains extend to cross-session evidence integration.}
QEMem achieves 81.3\% and 83.4\% accuracy on cross-session questions, improving over the strongest included baseline by 9.3 and 11.7 points.
For cross-session list/count questions, the corresponding gains are 6.4 and 10.2 points.
These results demonstrate QEMem's effectiveness on questions requiring complementary evidence from multiple turns and sessions.
Additional analyses of evidence cardinality, temporal dispersion, and statistical robustness are provided in Appendix~\ref{app:rq5_complexity}.

\begin{table*}[t]
\centering
\caption{
\textbf{LLM-judge accuracy (\%) across evidence-complexity groups on LoCoMo.}
Overlapping groups are defined by supporting turns and sessions among 1,535 questions with valid evidence annotations.
$\Delta$ is QEMem's gain over the strongest included baseline per group (percentage points).
Bold/underlined values mark the best/second-best accuracy.
}
\label{tab:rq5_complexity}
\setlength{\tabcolsep}{3pt}
\renewcommand{\arraystretch}{1.0}
\small

\noindent\textbf{(a) Claude Haiku 4.5}
\vspace{3pt}

\begin{tabular*}{\textwidth}{@{\extracolsep{\fill}}lcccccc@{}}
\toprule
\textbf{Evidence Complexity}
& $\boldsymbol{n}$
& \textbf{NEMORI}
& \textbf{SimpleMem}
& \textbf{MemR$^3$}
& \textbf{QEMem}
& $\boldsymbol{\Delta}$ \textbf{vs. Best} \\
\midrule
Single evidence turn
& 1122 & 79.9 & 80.8 & \underline{86.8}
& \textbf{92.1} & +5.3 \\
$\geq$3 evidence turns
& 184 & 63.6 & \underline{75.0} & 53.8
& \textbf{81.0} & +6.0 \\
Cross-session evidence
& 332 & 63.3 & \underline{72.0} & 59.0
& \textbf{81.3} & +9.3 \\
Cross-session list/count
& 235 & 59.6 & \underline{74.0} & 54.5
& \textbf{80.4} & +6.4 \\
\bottomrule
\end{tabular*}

\vspace{5pt}
\noindent\textbf{(b) Claude Opus 4.5}

\vspace{3pt}
\begin{tabular*}{\textwidth}{@{\extracolsep{\fill}}lcccccc@{}}
\toprule
\textbf{Evidence Complexity}
& $\boldsymbol{n}$
& \textbf{NEMORI}
& \textbf{SimpleMem}
& \textbf{MemR$^3$}
& \textbf{QEMem}
& $\boldsymbol{\Delta}$ \textbf{vs. Best} \\
\midrule
Single evidence turn
& 1122 & 84.8 & 83.8 & \underline{90.1}
& \textbf{94.0} & +3.9 \\
$\geq$3 evidence turns
& 184 & 64.1 & \underline{73.9} & 59.8
& \textbf{82.1} & +8.2 \\
Cross-session evidence
& 332 & 66.3 & \underline{71.7} & 63.3
& \textbf{83.4} & +11.7 \\
Cross-session list/count
& 235 & 60.9 & \underline{72.8} & 57.4
& \textbf{83.0} & +10.2 \\
\bottomrule
\end{tabular*}
\end{table*}

\begin{figure*}[t]
    \centering
    \includegraphics[width=\textwidth]{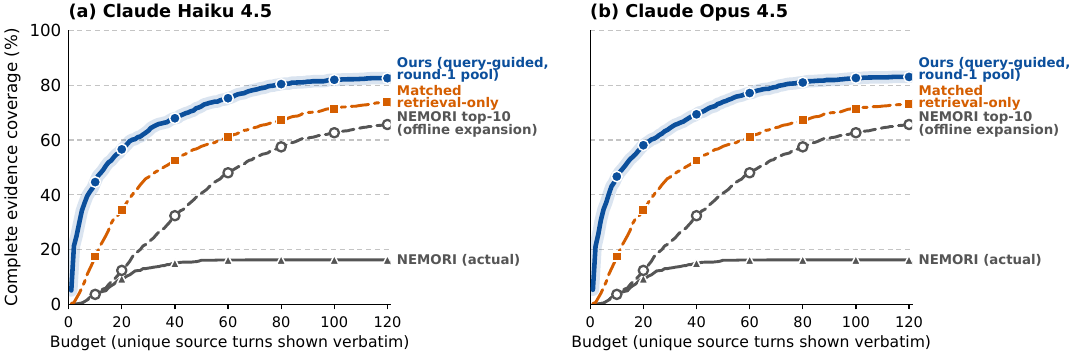}
\caption{
    \textbf{Complete evidence coverage under source-turn budgets.}
    \textbf{(a)} Haiku 4.5.
    \textbf{(b)} Opus 4.5.
    Curves measure coverage of all annotated supporting turns in offline retrieval prefixes, preserving retrieval order and candidate boundaries.
    Budgets count unique original turns exposed verbatim.
    QEMem uses its initial retrieval pool.
    NEMORI (actual) exposes source turns from its top-two episodes; NEMORI top-10 hypothetically expands all ten retrieved episodes into their associated source turns.
    Results cover 1,531 questions with fully mappable evidence annotations.
    }
    \label{fig:rq6_budget}
\end{figure*}

\subsection{RQ6: Evidence Coverage under Source-Turn Budgets}
\label{sec:rq6_budget}

We measure complete evidence coverage over prefixes of stored retrieval results, preserving retrieval order and limiting the number of unique source turns exposed.
Figure~\ref{fig:rq6_budget} evaluates QEMem's initial multi-granular retrieval pool before evidence reification and refinement.
\textbf{QEMem achieves higher evidence coverage under the same source-turn budgets.}
At 20 turns, coverage reaches 56.6\% with Haiku and 58.1\% with Opus, compared with 34.5\% for matched retrieval-only, yielding gains of 22.1 and 23.6 percentage points.
Across budgets of 10--120 turns, the normalized area under the coverage curve improves by 14.5 and 15.7 points, respectively.
QEMem also outperforms the hypothetical expansion of all retrieved NEMORI episodes into their associated source turns.
These results show that QEMem's initial retrieval provides more complete annotated evidence within limited source-turn prefixes.
Token-budget results, statistical robustness checks, and query-planning analyses are provided in Appendix~\ref{app:rq6_budget}.

\section{Conclusion}
\label{sec:conclusion}

We present QEMem, a framework for addressing the relevance--sufficiency gap in long-term agent memory through query-specific evidence construction.
QEMem combines multi-granular evidence discovery, source-grounded ViewCards, and targeted refinement over existing memory stores.
Experiments on LoCoMo show improved answer accuracy and evidence coverage across two LLM backbones, with strong performance on questions requiring cross-session evidence integration.
These results highlight query-time evidence construction as an effective approach to improving memory utilization while making supporting sources explicit.

\clearpage


\section*{Limitations}

Our evaluation focuses on LoCoMo and two Claude backbones; broader validation across benchmarks and model families remains future work. Evidence coverage relies on benchmark annotations, while grounding is assessed through automated checks and sampled LLM-based audits. QEMem also trades additional query-time computation for structured evidence and targeted refinement, motivating further work on reducing this overhead while preserving its benefits.
\bibliography{custom}
\clearpage
\appendix
\section*{Appendix}

\section{Related Work}
\label{app:related_work}

\subsection{Long-Term Memory for LLM Agents}
\label{subsec:rw_memory_systems}

Recent advances in large language models have enabled increasingly capable agentic systems that perform complex tasks through reasoning, interaction, and collaboration~\citep{chang2025survey,chen2026unitymas,yang2025toward,yang2026auditing,li2026fortis}. Multi-agent frameworks further coordinate specialized agents to improve problem solving and decision making~\citep{yang2026adaptive}, with applications spanning code generation~\citep{ping2026coevo,ping2026poet}, ranking and search~\citep{chen2025tourrank,zhang2026oases}, and reinforcement learning for agentic reasoning~\citep{yang2026tournament,zhang2026oases,ping2025verimoa}. As these systems evolve toward longer-horizon and more autonomous interactions, their effectiveness increasingly depends on the ability to accumulate experience, preserve relevant context, and reuse previously acquired information. This growing need motivates long-term memory as a fundamental component of LLM-based agents~\citep{park2023generativeagentsinteractivesimulacra,hu2025memory}.

Existing research has explored diverse mechanisms for persistent memory storage, organization, and retrieval in LLM agents~\citep{du2026memoryautonomousllmagentsmechanisms,qin2026memory,ping2026rem}. Early approaches primarily address how conversational histories can be retained and accessed over time. MemGPT~\citep{packer2024memgptllmsoperatingsystems} introduces hierarchical memory management inspired by operating systems, while MemoryBank~\citep{zhong2024memorybank} maintains persistent user-related information for personalized interactions. MemoryOS~\citep{kang2025memory} further organizes memory into multiple components to support long-term conversational continuity. 

Subsequent research explores how memory representations can become more compact, structured, and informative. Mem0~\citep{chhikara2025mem0} extracts persistent facts from dialogue, while A-Mem~\citep{xu2025mem} develops interconnected memory notes that evolve as new experiences arrive. Zep~\citep{rasmussen2025zep} organizes conversational knowledge through temporal graph structures, supporting relationships and updates across events. LightMem~\citep{fang2025lightmem} emphasizes efficient memory consolidation through a lightweight architecture, whereas SimpleMem~\citep{liu2026simplemem} constructs self-contained memories with complementary retrieval views. MemGAS~\citep{xu2026single} further organizes memories at multiple granularities and adaptively selects retrieval representations based on query characteristics. Collectively, these methods demonstrate the importance of memory organization and consolidation for effective long-term information access.

Beyond memory representation, recent systems increasingly adapt memory operations to the agent's changing information needs. NEMORI~\citep{nan2025nemori} connects memory formation with predictability and prediction error, while AgeMem~\citep{yu2026agentic} and MemAgent~\citep{yu2025memagent} explore learned memory-management behaviors. These developments improve how agents retain, update, and retrieve past information. However, the quality of stored memories and their individual relevance to a query do not necessarily determine whether the retrieved information collectively supports a complete answer. QEMem addresses this complementary problem at query time, treating the construction of a sufficient, source-grounded evidence set as the central objective rather than introducing another memory storage architecture.

\subsection{Retrieval and Evidence Reasoning}
\label{subsec:rw_context_retrieval}

Effective memory utilization is closely related to retrieval-augmented generation and long-context reasoning. Retrieval-augmented generation~\citep{lewis2020retrieval} provides external information to a language model, while dense retrieval~\citep{karpukhin2020densepassageretrievalopendomain} and lexical retrieval~\citep{10.1561/1500000019} offer complementary mechanisms for identifying relevant content. Alternatively, long-context architectures allow models to process larger amounts of information directly~\citep{beltagy2020longformer,zaheer2020big,dai-etal-2019-transformer}. Nevertheless, access to additional context does not guarantee effective evidence utilization, as illustrated by the lost-in-the-middle phenomenon~\citep{liu2024lost}. Methods such as LLMLingua~\citep{jiang2023llmlingua} and ReadAgent~\citep{lee2024human} reduce context demands through compression or structured reading, while GraphRAG~\citep{edge2024local} and LightRAG~\citep{guo2024lightrag} introduce graph-based organization to facilitate information retrieval and integration. Joren et al.\citet{joren2025sufficient} distinguish insufficient context from failures to use available information in RAG.
CRAG~\citep{yan2024corrective} evaluates retrieval quality to guide corrective retrieval actions.
Complementary work examines source attribution: RARR~\citep{gao2023rarr} retrieves evidence to revise unsupported generated content, while ALCE~\citep{gao2023enabling} evaluates answer correctness and citation quality.
QEMem brings these concerns to long-term agent memory by constructing source-linked evidence sets and refining them to address unresolved information needs before answering. These approaches improve information accessibility, but retrieving relevant content remains distinct from establishing whether the available evidence is sufficient for a particular question.

A complementary direction incorporates reasoning and feedback into retrieval decisions. ReAct~\citep{yao2022react} interleaves reasoning with external actions, while Self-RAG~\citep{asai2023self} and active retrieval~\citep{jiang2023active} allow retrieval behavior to respond to evolving generation needs. More recently, Search-R1~\citep{jin2025searchr1trainingllmsreason} explores reinforcement learning for multi-step search and reasoning. In long-term conversational memory, MemR$^3$~\citep{du2025memr} coordinates retrieval, reflection, and answering through an iterative reasoning process. REMem~\citep{shu2026remem} combines episodic memory graphs with agentic retrieval to support recollection and reasoning over past events, while MRAgent~\citep{ji2026memory} employs associative graph structures and active memory reconstruction. E-mem~\citep{wang2026mem} further investigates episodic context reconstruction through multi-agent reasoning within retrieved memory segments. Together, these approaches move beyond single-pass retrieval toward more adaptive and context-aware memory reasoning.

Despite this progress, long-term conversational histories present a particular challenge: relevant evidence may be distributed across different sessions, represented at different levels of abstraction, or embedded within memories that contain only part of the information required to answer a query~\citep{maharana2024evaluating,jia-etal-2025-evaluating}. Consequently, retrieving individually relevant memories does not necessarily produce a collectively sufficient evidence set. Existing approaches emphasize memory representation, retrieval control, or iterative reasoning, whereas QEMem explicitly centers on \emph{query-specific evidence-set construction}. It combines query-guided evidence discovery with source-grounded evidence reification, making supporting claims and their provenance explicit, and uses sufficiency feedback to refine the evidence set when necessary. This perspective shifts the objective from retrieving relevant memory units to assembling complementary, traceable evidence that collectively supports the information requirements of the query.

\section{Experiment}
\label{app:exp}

\subsection{Experimental Setup}
\label{app:experimental_setup}

\paragraph{Dataset.}
We evaluate QEMem on \textbf{LoCoMo}, a benchmark for long-term conversational memory comprising 10 extended, multi-session conversations. Following the standard evaluation protocol, we consider 1,540 questions across four categories: Multi-hop (282), Temporal (321), Open-domain (96), and Single-hop (841), excluding adversarial questions. LoCoMo also provides annotated supporting dialogue turns, which we use exclusively for post-hoc evidence analysis and never expose during inference.

\paragraph{Baselines.}
We compare QEMem against seven representative memory systems: \textbf{NEMORI}, \textbf{Mem0}, \textbf{A-Mem}, \textbf{LightMem}, \textbf{SimpleMem}, \textbf{MemR$^3$}, and \textbf{MemGAS}. These methods cover episodic and semantic memory organization, memory extraction and consolidation, compression-based memory management, and iterative or structured memory retrieval. We use their official implementations or adaptations of the released code, following the recommended LoCoMo configurations whenever applicable. All baselines are evaluated on the same questions under consistent backbone and evaluation settings. In particular, QEMem and NEMORI share an identical preconstructed memory store, enabling a controlled comparison of their query-time memory utilization mechanisms.

\paragraph{Implementation Details.}
We evaluate two LLM backbones, \textbf{Claude Haiku 4.5} and \textbf{Claude Opus 4.5}, accessed through Amazon Bedrock. Memory construction is performed once using Opus 4.5, and the resulting memories are reused across both backbone settings. QEMem operates on NEMORI's episodic, semantic, and original dialogue memories, employing dense retrieval and BM25 with reciprocal rank fusion for multi-granular evidence discovery. We use \texttt{Qwen3-Embedding-4B} ~\citep{zhang2025qwen3embeddingadvancingtext} as the embedding model for QEMem and compatible baselines, retaining the original embedding models when required by baseline-specific mechanisms. All LLM inference uses a temperature of zero. Following the NEMORI evaluation protocol, we report LLM-judge accuracy as the primary metric, using a fixed Opus 4.5 judge and the same evaluation prompt across all methods. We additionally conduct paired statistical comparisons and evidence-level analyses using existing predictions and execution traces, without exposing reference answers or supporting-evidence annotations during inference.

\subsection{Additional Evidence Sufficiency Analysis}
\label{app:rq2_haiku}

To examine whether the relevance--sufficiency gap persists across answer models, we extend the RQ2 analysis to Claude Haiku 4.5, as summarized in Table~\ref{tab:rq2_haiku}. The results closely reproduce the evidence-recovery patterns observed with Opus 4.5. QEMem increases complete retrieval coverage from 68.8\% to 93.5\%, while complete source visibility rises from 16.2\% to 91.6\%. Moreover, ViewCard sources cover all annotated supporting turns for 83.3\% of questions, including direct turn citations and provenance links through cited episodes. The improvement is particularly substantial for multi-hop questions, where complete source visibility increases from 1.8\% to 72.8\%. More importantly, the relationship between evidence availability and answer accuracy remains consistent across backbones. When NEMORI retrieves records containing all supporting evidence, QEMem provides a modest 2.1-point improvement. In contrast, the gains increase to 22.0 and 47.5 percentage points when baseline retrieval coverage is partial or missing, respectively. Although these two groups represent only 31.2\% of questions, they account for 89.1\% of QEMem's net accuracy improvement. Together with the Opus results, these findings suggest that incomplete evidence recovery is a persistent bottleneck across model capabilities, and that constructing query-specific evidence sets is especially beneficial when conventional retrieval fails to recover the necessary supporting information.

\paragraph{Coverage computation.}
Visibility is computed from the final turn IDs used to assemble the answer prompt, including card-derived source context.
Card-source coverage uses resolvable turn and episode citations regardless of quotation-verification status.
Semantic-memory provenance is excluded from these coverage metrics.
Turn IDs sharing the same timestamp, speaker role, and content are treated as aliases.

\begin{table*}[t]
\centering
\caption{
Additional RQ2 analysis with Claude Haiku 4.5.
\textbf{(a)} Complete coverage of annotated supporting evidence across question categories.
R denotes retrieval-record coverage, V denotes original source-turn visibility in the answer prompt, and C denotes ViewCard citation coverage.
\textbf{(b)} Answer accuracy conditioned on NEMORI's retrieval coverage.
All values are percentages unless otherwise stated.
Results are computed on 1,531 questions with fully mappable evidence annotations.
}
\label{tab:rq2_haiku}
\setlength{\tabcolsep}{7pt}
\renewcommand{\arraystretch}{1.10}
\small

\textbf{(a) Complete Evidence Coverage}
\vspace{3pt}

\begin{tabular}{lccccc}
\toprule
\textbf{Method / Metric}
& \textbf{Multi-Hop}
& \textbf{Temporal}
& \textbf{Open-Domain}
& \textbf{Single-Hop}
& \textbf{Overall} \\
\midrule
NEMORI -- R
& 31.5 & 75.6 & 38.0 & 82.0 & 68.8 \\
NEMORI -- V
& 1.8 & 19.7 & 8.7 & 20.5 & 16.2 \\
\midrule
QEMem -- R
& \textbf{78.8} & \textbf{97.8} & \textbf{70.7}
& \textbf{99.3} & \textbf{93.5} \\
QEMem -- V
& 72.8 & 95.9 & 69.6 & 98.7 & 91.6 \\
QEMem -- C
& 49.5 & 90.9 & 59.8 & 94.2 & 83.3 \\
\bottomrule
\end{tabular}

\vspace{9pt}

\textbf{(b) Accuracy Conditioned on NEMORI Retrieval Coverage}
\vspace{3pt}

\begin{tabular}{lccccc}
\toprule
\textbf{Coverage}
& \textbf{Count}
& \textbf{NEMORI}
& \textbf{QEMem}
& $\boldsymbol{\Delta}$ \textbf{(pp)}
& \textbf{95\% CI (pp)} \\
\midrule
Complete
& 1054 & 90.4 & 92.5 & +2.1 & [+0.7, +3.6] \\
Partial
& 182 & 59.9 & 81.9 & \textbf{+22.0} & [+15.4, +29.1] \\
Missing
& 295 & 36.6 & 84.1 & \textbf{+47.5} & [+41.4, +53.6] \\
\midrule
Overall
& 1531 & 76.4 & 89.6 & \textbf{+13.2} & [+11.2, +15.2] \\
\bottomrule
\end{tabular}

\end{table*}

\subsection{Additional Analysis of Source Grounding and Evidential Reliability}
\label{app:rq4_grounding}

\paragraph{Claim--Source Semantic Consistency.}
While quotation fidelity establishes whether cited text can be recovered from the referenced memory records, it does not establish whether the corresponding evidence claims are logically supported. To evaluate this distinction, we conduct an independent claim--source consistency audit over 288 stratified samples (144 ViewCards per backbone), sampled across question categories and provenance types. An Opus 4.5 judge evaluates each claim using only its cited source content and timestamps, without access to the question, gold answer, or answer-correctness labels. As shown in Table~\ref{tab:rq4_semantic}, 97.2\% of Haiku-generated claims and 98.6\% of Opus-generated claims are fully or partially supported by their cited evidence. Under the stricter criterion of complete semantic entailment, the corresponding population-weighted estimates are 59.6\% (95\% CI: [49.5, 70.2]) and 67.0\% (95\% CI: [57.3, 76.5]). Unsupported or contradicted claims account for only 1.1\% and 1.3\%, respectively, under population weighting. The distinction between full and partial support is important: many partially supported claims correctly recover the central fact but introduce additional details not explicitly established by the source, particularly temporal or entity-level attributions. Thus, high quotation fidelity should be interpreted as evidence traceability rather than a guarantee of complete factual entailment.

\begin{table*}[t]
\centering
\caption{
\textbf{Claim--source semantic consistency of sampled ViewCards.}
Results are based on 288 stratified samples evaluated by Claude Opus 4.5. The percentages below are unweighted sample proportions. Fully or partially supported claims include both strict entailment and claims containing additional details not fully established by the cited sources. Population-weighted estimates are reported in the text.
}
\label{tab:rq4_semantic}
\setlength{\tabcolsep}{7pt}
\renewcommand{\arraystretch}{1.12}
\small
\begin{tabular}{lcccccc}
\toprule
\textbf{Backbone}
& \textbf{Samples}
& \textbf{Fully Supported}
& \textbf{Partially Supported}
& \textbf{Unsupported}
& \textbf{Contradicted}
& \textbf{Full + Partial} \\
\midrule
Haiku 4.5
& 144 & 57.6 & 39.6 & 0.7 & 2.1 & \textbf{97.2} \\
Opus 4.5
& 144 & 66.0 & 32.6 & 0.7 & 0.7 & \textbf{98.6} \\
\bottomrule
\end{tabular}
\end{table*}

\paragraph{Temporal Grounding Reliability.}
We further audit temporal grounding by comparing ViewCard time fields against mechanically checkable temporal relations derived from their cited sources and conversation timestamps. As summarized in Table~\ref{tab:rq4_temporal}, 93.9\% of Haiku-generated and 96.1\% of Opus-generated time fields agree with the corresponding deterministic verification rules. Importantly, these results are not solely attributable to the pipeline's deterministic time normalizer: when restricting the evaluation to time fields originally produced by the LLM, consistency remains 93.3\% and 95.7\%, respectively. However, temporal normalization remains less reliable for particular relative expressions. For instance, consistency on ``yesterday'' or ``last night'' is 62.4\% for Haiku and 80.7\% for Opus, while references to ``last year'' achieve 60.9\% and 75.4\%. These findings indicate that explicit temporal anchoring enables useful consistency checks, while relative-time interpretation remains a source of potential errors. Since the audit measures agreement with deterministic rules rather than independently verified event dates, the reported consistency rates should not be interpreted as comprehensive temporal factual accuracy.

\begin{table*}[t]
\centering
\caption{
\textbf{Temporal-grounding consistency of QEMem ViewCards.}
The audit covers cards with explicit time fields and reports agreement with deterministic source-timestamp-based rules. LLM-written time fields exclude values rewritten by the pipeline's deterministic normalizer. Relative-time results are reported over applicable cases only.
}
\label{tab:rq4_temporal}
\setlength{\tabcolsep}{5pt}
\renewcommand{\arraystretch}{1.12}
\footnotesize
\begin{tabular}{lccccc}
\toprule
\textbf{Backbone}
& \textbf{Time Fields}
& \shortstack{\textbf{Mechanically}\\\textbf{Verified}}
& \shortstack{\textbf{LLM-Written}\\\textbf{Time Consistency}}
& \shortstack{\textbf{Yesterday /}\\\textbf{Last Night}}
& \textbf{Last Year} \\
\midrule
Haiku 4.5
& 9,541
& 5,398/5,746 (93.9\%)
& 4,857/5,205 (93.3\%)
& 121/194 (62.4\%)
& 39/64 (60.9\%) \\
Opus 4.5
& 9,598
& 5,822/6,060 (96.1\%)
& 5,287/5,525 (95.7\%)
& 176/218 (80.7\%)
& 49/65 (75.4\%) \\
\bottomrule
\end{tabular}
\end{table*}

\paragraph{Failure Modes and Limitations.}
The audit reveals several limitations of source-grounded evidence construction. First, a small fraction of Haiku-generated cards (1.6\%) contain unresolved provenance references, primarily because the model refers to evidence from previous refinement rounds without providing valid source identifiers; no such failures are observed for Opus. Second, citations to episodic memories frequently reproduce text from the episode summaries rather than verbatim original dialogue. Such citations remain traceable to memory records, but their fidelity to the underlying conversations cannot be established through quotation matching alone. Third, semantic inconsistencies most commonly arise from over-specific temporal claims, incorrect attribution of information to speakers or events, and the merging of facts from different sources. Finally, the semantic audit relies on a limited stratified sample and a single LLM judge, while the temporal audit covers only mechanically interpretable expressions. Together, these observations clarify the scope of QEMem's grounding guarantees: its evidence representations provide explicit and verifiable provenance, but source traceability and quotation fidelity do not imply complete semantic correctness.

\subsection{Performance--Cost Tradeoffs and Adaptive Computation}
\label{app:rq5_efficiency}

\paragraph{Evaluation Protocol.}
We analyze the query-time inference cost of QEMem and its variants on all 1,540 LoCoMo questions using Claude Haiku 4.5 and Opus 4.5. We account for all query-time LLM calls, including query planning, evidence construction and refinement, and final answer generation. Offline evaluation and one-time memory construction are excluded from per-query costs. All configurations share the same preconstructed memory store, isolating the cost of query-time memory utilization. Cost estimates follow the pricing used in our experiments: \$1.1/\$5.5 per million input/output tokens for Haiku 4.5 and \$5/\$25 for Opus 4.5. Full QEMem and retrieval-only costs are derived from recorded token usage, while costs for variants reusing intermediate stages are reconstructed from saved traces and measured answer-generation usage.

\paragraph{Ablation Implementation.}
Query-Guided Retrieval answers directly from the initial candidate pool. The variant without refinement additionally constructs ViewCards, verifies quotations, normalizes temporal fields, and assembles cited-source context. Turn citations include up to two neighboring turns on each side within the same session, while episode citations expose the episode's first twelve original turns. These source expansions introduce no follow-up retrieval. The direct-answer variant removes card-specific prompt sections while retaining the same substantive answering rules and generation settings.

\paragraph{Trace Reconstruction and Sensitivity.}
For questions that triggered refinement, initial semantic and turn pools were reconstructed through retrieval replay, with a median Jaccard similarity of approximately 0.97 in the reconstruction check. For questions without refinement, the saved pool is the initial pool, and the no-refinement variant reuses Full QEMem's answer. Among the 131 Opus refinement questions, 113 have unambiguous first-round card boundaries. On this subset, refinement increases accuracy from 60.2\% to 71.7\% ($+11.5$ percentage points; 95\% CI: $[+5.3,+18.6]$), consistent with the improvement observed across all refinement-triggered questions.

\begin{figure*}[t]
    \centering
    \includegraphics[width=\textwidth]{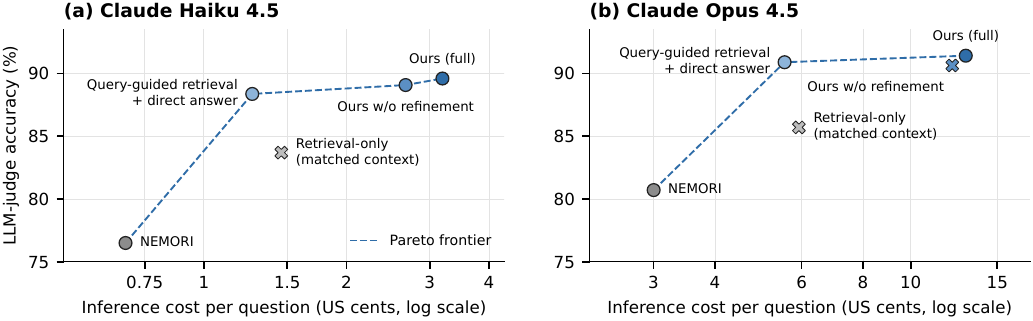}
    \caption{
    \textbf{Accuracy--cost tradeoffs on LoCoMo.}
    Query-time inference cost versus LLM-judge accuracy for
    \textbf{(a)} Claude Haiku 4.5 and
    \textbf{(b)} Claude Opus 4.5.
    Dashed lines indicate the Pareto frontier within each backbone.
    Query-guided retrieval with direct answering improves both
    accuracy and cost relative to matched-context retrieval.
    Full QEMem achieves the highest observed accuracy while providing
    structured evidence representations and selective refinement.
    Costs include all query-time LLM calls but exclude shared
    memory construction and offline evaluation.
    }
    \label{fig:rq5_efficiency}
\end{figure*}

\paragraph{Accuracy--Cost Tradeoffs.}
Figure~\ref{fig:rq5_efficiency} and Table~\ref{tab:rq5_efficiency} illustrate how different evidence-construction configurations balance inference cost and answer quality. \textbf{Query-guided multi-granular retrieval provides a particularly favorable accuracy--cost tradeoff.} Compared with matched-context retrieval, direct answering from the query-guided candidate pool improves accuracy by 4.7 percentage points with Haiku 4.5 and 5.2 points with Opus 4.5, while reducing inference cost from 1.46 to 1.27 cents and from 5.93 to 5.54 cents per question, respectively. These results demonstrate that more effective evidence discovery can improve answer quality while reducing redundant context processing. \textbf{Full QEMem achieves the highest overall accuracy} under both backbones, reaching 89.6\% with Haiku and 91.4\% with Opus. Its additional computation supports structured evidence construction, explicit source grounding, and targeted refinement, increasing complete evidence visibility to 91.6\% and 91.7\%, respectively. Meanwhile, direct answering from query-guided retrieval offers a lightweight alternative, achieving 88.4\% and 90.9\% accuracy at approximately 40--43\% of Full QEMem's query-time cost. These operating points illustrate the flexibility of query-specific evidence construction, from efficient evidence discovery to more comprehensive, source-grounded reasoning.

\begin{table*}[t]
\centering
\caption{
\textbf{Query-time efficiency comparison on LoCoMo.}
Accuracy and complete evidence visibility are percentages.
Input and output tokens are average counts per question.
Cost is reported in US cents per question.
Visibility measures complete coverage of annotated supporting
turns in the final answer context, evaluated on 1,531 questions
with fully mappable evidence annotations.
}
\label{tab:rq5_efficiency}
\setlength{\tabcolsep}{5pt}
\renewcommand{\arraystretch}{1.12}
\footnotesize

\begin{tabular*}{\textwidth}{@{\extracolsep{\fill}}lcccccc@{}}
\toprule
\textbf{Method}
& \textbf{Accuracy}
& \textbf{Calls}
& \textbf{Input Tokens}
& \textbf{Output Tokens}
& \textbf{Cost (cents)}
& \textbf{Visibility} \\
\midrule
\multicolumn{7}{l}{\textbf{(a) Claude Haiku 4.5}} \\
\midrule
NEMORI
& 76.5 & 1.00 & 5,186 & 205 & 0.68 & 16.2 \\
Matched-Budget Retrieval
& 83.7 & 1.04 & 12,119 & 224 & 1.46 & 77.0 \\
Query-Guided Retrieval (Direct Answer)
& 88.4 & 2.02 & 9,784 & 344 & 1.27 & 82.6 \\
QEMem w/o Refinement
& 89.1 & 3.01 & 19,396 & 971 & 2.67 & 89.3 \\
\textbf{Full QEMem}
& \textbf{89.6} & 3.34 & 23,003 & 1,199
& 3.19 & \textbf{91.6} \\
\midrule
\multicolumn{7}{l}{\textbf{(b) Claude Opus 4.5}} \\
\midrule
NEMORI
& 80.7 & 1.00 & 5,186 & 164 & 3.00 & 16.2 \\
Matched-Budget Retrieval
& 85.7 & 1.03 & 10,776 & 215 & 5.93 & 75.0 \\
Query-Guided Retrieval (Direct Answer)
& 90.9 & 2.01 & 9,422 & 332 & 5.54 & 83.1 \\
QEMem w/o Refinement
& 90.6 & 3.02 & 19,255 & 1,008 & 12.15 & 90.7 \\
\textbf{Full QEMem}
& \textbf{91.4} & 3.13 & 20,499 & 1,077
& 12.94 & \textbf{91.7} \\
\bottomrule
\end{tabular*}

\vspace{3pt}
\begin{minipage}{0.98\textwidth}
\footnotesize
\textit{Note:} Matched-Budget Retrieval matches the final
answer-context token budget, rather than total inference
computation. Query-Guided Retrieval (Direct Answer) retains
query planning and multi-granular retrieval but omits
ViewCard construction and refinement. Cost estimates for
variants reusing intermediate stages include reconstructed
usage from those stages.
\end{minipage}
\end{table*}

\paragraph{Marginal Value of Selective Refinement.}
We further investigate the marginal benefit of sufficiency-guided refinement and its adaptive allocation of computation. Under Opus 4.5, refinement is triggered for only 8.5\% of questions, concentrating additional retrieval on cases with unresolved information needs. Within this subset, accuracy increases from 61.8\% to 71.0\%, yielding a \textbf{9.2-point improvement} ($p=0.0075$). Averaged over the entire benchmark, refinement adds approximately 0.79 cents per query while improving overall accuracy by 0.8 points. Under Haiku 4.5, refinement is triggered for 21.5\% of questions and yields a further 2.4-point improvement within the selected subset. Importantly, refinement is more frequently activated when the initial evidence is incomplete: with Opus, questions lacking complete annotated evidence trigger refinement at a rate of 21.1\%, compared with 7.3\% when all supporting evidence is already visible. This pattern demonstrates that sufficiency feedback is responsive to evidence availability, directing additional computation toward queries more likely to require complementary information.

\paragraph{Cost Distribution and Limitations.}
Full QEMem requires an average of 3.34 LLM calls per question under Haiku 4.5 and 3.13 under Opus 4.5. Across the two backbones, approximately 3--4\% of query-time cost is attributable to planning, 51--52\% to evidence construction and refinement, and 45--46\% to final answering. The modest planning overhead contrasts with the greater computational investment in evidence construction, which provides structured provenance, improves source visibility, and enables targeted refinement. Consequently, query-guided retrieval with direct answering represents an economical operating point, while Full QEMem achieves the highest observed accuracy and offers explicitly traceable evidence representations. These tradeoffs reflect the distinct roles of evidence discovery and evidence-set construction rather than a uniform improvement in computational efficiency. The analysis is limited to the evaluated configurations, and certain ablation costs are reconstructed from intermediate traces. Finally, the shared memory store incurs a one-time construction cost of approximately \$60.78 across the ten LoCoMo conversations, which is excluded from the query-time comparisons.

\subsection{Additional Analysis of Evidence Complexity and Compositional Reasoning}
\label{app:rq5_complexity}

\paragraph{Evidence Complexity Definitions.}
We characterize evidence complexity using the annotated supporting dialogue turns provided by LoCoMo. Specifically, we consider three complementary dimensions: (i) evidence cardinality, measured by the number of distinct annotated supporting turns; (ii) evidence dispersion, measured by the number of distinct conversation sessions containing these turns; and (iii) temporal dispersion, measured by the elapsed time between the earliest and latest supporting sessions. These characteristics are computed exclusively from benchmark annotations and conversation metadata, without using model predictions or retrieval outcomes. Of the 1,540 benchmark questions, 1,535 contain at least one valid supporting-turn annotation and are included in this analysis. Four questions with partially malformed annotations are retained using their valid supporting identifiers. Importantly, the number of annotated supporting turns is an observable proxy for evidence complexity, rather than a definitive measure of the minimum number of facts required to answer a question.

\begin{figure*}[t]
    \centering
    \includegraphics[width=\textwidth]{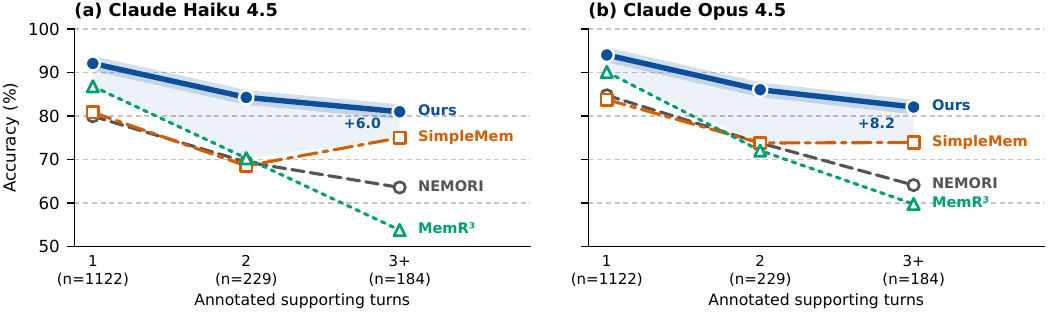}
    \caption{
    \textbf{Answer accuracy across evidence cardinality on LoCoMo.}
    Results are grouped by the number of annotated supporting dialogue turns:
    one turn ($n=1{,}122$), two turns ($n=229$), and three or more turns ($n=184$).
    \textbf{(a)} Claude Haiku 4.5.
    \textbf{(b)} Claude Opus 4.5.
    QEMem maintains the highest absolute accuracy among the evaluated methods across all complexity groups, while NEMORI and MemR$^3$ exhibit substantially greater performance degradation as evidence cardinality increases.
    }
    \label{fig:rq5_cardinality}
\end{figure*}

\paragraph{Performance across Evidence Cardinality.}
Figure~\ref{fig:rq5_cardinality} examines how answer accuracy varies with the number of annotated supporting turns. Several memory systems exhibit substantial performance degradation as evidence requirements become more complex. Under Haiku 4.5, NEMORI decreases from 79.9\% to 63.6\%, while MemR$^3$ decreases from 86.8\% to 53.8\% when moving from one supporting turn to three or more. In contrast, QEMem maintains 81.0\% accuracy on questions requiring at least three supporting turns, compared with 92.1\% on single-turn questions. Similar patterns emerge under Opus 4.5, where QEMem achieves 82.1\% on the most evidence-intensive group, compared with 64.1\% for NEMORI and 59.8\% for MemR$^3$. Although SimpleMem exhibits a smaller performance decline as evidence cardinality increases, QEMem consistently maintains higher absolute accuracy across all evaluated groups. These results suggest that query-specific evidence construction is particularly beneficial when answering requires integrating complementary information from multiple sources.

\paragraph{Temporal Dispersion of Supporting Evidence.}
We further examine questions requiring multiple supporting turns distributed over different time periods. Table~\ref{tab:rq5_temporal} reports accuracy across four temporal-span intervals, measured using the source conversation timestamps. QEMem achieves the highest accuracy among the evaluated methods across all temporal-span groups. On questions whose supporting evidence spans more than 180 days, QEMem obtains 71.2\% accuracy with Haiku 4.5 and 74.2\% with Opus 4.5, exceeding NEMORI by 19.7 and 24.2 percentage points, respectively. The corresponding improvements over MemR$^3$ are 22.7 points under both backbones. These observations further support QEMem's ability to integrate temporally dispersed evidence. However, the longest-span subset contains only 66 questions, and temporal span is strongly correlated with multi-hop question categories. Accordingly, these findings should be interpreted as descriptive performance patterns rather than independent effects of temporal distance.

\begin{table*}[t]
\centering
\caption{
\textbf{Accuracy across temporal dispersion of supporting evidence.}
Temporal span is measured between the earliest and latest annotated supporting sessions for questions with multiple supporting turns.
All results are LLM-judge accuracy percentages.
}
\label{tab:rq5_temporal}
\setlength{\tabcolsep}{7pt}
\renewcommand{\arraystretch}{1.10}
\footnotesize

\begin{tabular*}{\textwidth}{@{\extracolsep{\fill}}lccccc@{}}
\toprule
\textbf{Temporal Span}
& $\boldsymbol{n}$
& \textbf{NEMORI}
& \textbf{SimpleMem}
& \textbf{MemR$^3$}
& \textbf{QEMem} \\
\midrule
\multicolumn{6}{l}{\textbf{(a) Claude Haiku 4.5}} \\
\midrule
Same day
& 81 & 81.5 & 69.1 & 79.0 & \textbf{88.9} \\
1--30 days
& 83 & 62.7 & 72.3 & 66.3 & \textbf{80.7} \\
31--180 days
& 183 & 67.8 & 73.2 & 59.6 & \textbf{85.2} \\
$>180$ days
& 66 & 51.5 & 68.2 & 48.5 & \textbf{71.2} \\
\midrule
\multicolumn{6}{l}{\textbf{(b) Claude Opus 4.5}} \\
\midrule
Same day
& 81 & 82.7 & 82.7 & 80.2 & \textbf{87.7} \\
1--30 days
& 83 & 68.7 & 74.7 & 68.7 & \textbf{83.1} \\
31--180 days
& 183 & 71.0 & 72.1 & 65.0 & \textbf{86.9} \\
$>180$ days
& 66 & 50.0 & 66.7 & 51.5 & \textbf{74.2} \\
\bottomrule
\end{tabular*}
\end{table*}

\paragraph{Robustness of Complexity-Dependent Improvements.}
To examine whether QEMem's relative advantage changes with evidence complexity, we measure the difference in accuracy gaps between high- and low-complexity groups. For a baseline $b$, we define
\[
\begin{aligned}
\Delta_{\mathrm{gap}}
={}& \left(\mathrm{Acc}_{\mathrm{QEMem}}
          -\mathrm{Acc}_{b}\right)_{\mathrm{high}} \\
 &- \left(\mathrm{Acc}_{\mathrm{QEMem}}
          -\mathrm{Acc}_{b}\right)_{\mathrm{low}}.
\end{aligned}
\]
A positive value indicates that QEMem's advantage over the baseline is larger in the high-complexity group. Table~\ref{tab:rq5_robustness} reports these gap differences with conversation-cluster bootstrap confidence intervals, using the ten LoCoMo conversations as resampling units. Compared with MemR$^3$, QEMem's relative advantage increases by 21.9 and 18.4 percentage points from one supporting turn to three or more under Haiku and Opus, respectively. When moving from single-session to cross-session evidence, the corresponding increases are 16.7 and 16.0 points. These improvements remain positive under leave-one-conversation-out analysis, indicating that the relative advantages are not driven by a single conversation. Compared with NEMORI, QEMem also exhibits larger advantages in high-complexity groups, particularly under Opus 4.5. By contrast, SimpleMem demonstrates relatively stable performance across complexity levels, and QEMem's relative advantage does not consistently increase against this baseline. Nevertheless, QEMem maintains higher absolute accuracy across the reported complexity groups.

\begin{table*}[t]
\centering
\caption{
\textbf{Change in QEMem's accuracy advantage under increasing evidence complexity.}
Values represent $\Delta_{\mathrm{gap}}$ in percentage points with 95\% conversation-cluster bootstrap confidence intervals.
Positive values indicate a larger advantage for QEMem in high-complexity groups.
}
\label{tab:rq5_robustness}
\setlength{\tabcolsep}{5pt}
\renewcommand{\arraystretch}{1.12}
\footnotesize

\begin{tabular*}{\textwidth}{@{\extracolsep{\fill}}llcc@{}}
\toprule
\textbf{Complexity Contrast}
& \textbf{Baseline}
& \textbf{Haiku 4.5}
& \textbf{Opus 4.5} \\
\midrule
\multicolumn{4}{l}{\textit{1 supporting turn $\rightarrow$ 3+ supporting turns}} \\
& NEMORI
& $+5.3\;[-4.4,\,+16.8]$
& $+8.7\;[+3.0,\,+15.5]$ \\
& SimpleMem
& $-5.3\;[-14.7,\,+4.2]$
& $-2.1\;[-7.8,\,+4.6]$ \\
& MemR$^3$
& $\mathbf{+21.9}\;[+11.1,\,+36.4]$
& $\mathbf{+18.4}\;[+11.9,\,+27.9]$ \\
\midrule
\multicolumn{4}{l}{\textit{Single-session $\rightarrow$ cross-session evidence}} \\
& NEMORI
& $+6.3\;[-0.4,\,+13.4]$
& $+8.2\;[+4.5,\,+11.6]$ \\
& SimpleMem
& $-2.5\;[-8.8,\,+4.6]$
& $+1.9\;[-3.0,\,+7.9]$ \\
& MemR$^3$
& $\mathbf{+16.7}\;[+9.9,\,+24.4]$
& $\mathbf{+16.0}\;[+11.8,\,+21.7]$ \\
\bottomrule
\end{tabular*}
\end{table*}

\paragraph{Limitations of Complexity-Based Analysis.}
Evidence complexity is not independent of question type in LoCoMo. In particular, 269 of the 332 cross-session questions belong to the multi-hop category, while most single-turn questions are single-hop or temporal questions. This correlation limits the extent to which performance differences can be attributed specifically to evidence cardinality or dispersion. Moreover, the annotations may contain redundant evidence or omit alternative supporting turns, and some temporal-dispersion groups contain relatively few questions. The robustness analysis accounts for conversation-level variation, but the benchmark contains only ten conversations. Therefore, the results support an association between distributed evidence requirements and QEMem's relative performance advantages, rather than a causal claim that evidence dispersion alone determines the observed improvements.

\subsection{Additional Analysis of Budgeted Evidence Recovery}
\label{app:rq6_budget}

\paragraph{Evaluation Protocol.}
We examine evidence-recovery efficiency using the stored retrieval outputs from QEMem, NEMORI, and the matched-budget retrieval-only baseline introduced in Section~\ref{sec:rq3}. For each question, we preserve the original retrieval ordering and progressively expose source turns under a fixed budget. A supporting turn is considered recovered only when its original dialogue content is explicitly exposed; episodic summaries and semantic memories are not treated as verbatim source evidence. We deduplicate source turns across retrieved records and adopt a strict budget rule: retrieval stops when adding the next candidate would exceed the specified budget. This rule preserves the atomicity of retrieved episodes and source-turn candidates. The primary evaluation covers 1,531 questions whose annotated supporting turns can be mapped exactly to the original memory store. Importantly, this is an offline prefix analysis of previously generated retrieval lists, not a rerun of the retrieval pipeline under alternative budgets.

\paragraph{Coverage under Source-Turn Budgets.}
Table~\ref{tab:rq6_budget} presents complete evidence coverage at representative retrieval budgets. QEMem's initial retrieval pool consistently achieves higher coverage than matched retrieval-only across both backbones. At 20 turns, the improvements are 22.1 and 23.6 percentage points for Haiku 4.5 and Opus 4.5, respectively. At 80 turns, the corresponding improvements remain 13.2 and 13.8 points. Across the full 10--120-turn interval, the normalized area under the coverage curve increases from 58.6\% to 73.1\% under Haiku and from 58.5\% to 74.1\% under Opus. These differences remain positive across all ten leave-one-conversation-out evaluations, with conversation-cluster bootstrap confidence intervals excluding zero. Furthermore, QEMem reaches 80\% complete coverage with approximately 78 and 73 source turns under Haiku and Opus, respectively, whereas neither matched retrieval-only nor NEMORI reaches this threshold within its available retrieval list.

\begin{table*}[t]
\centering
\caption{
\textbf{Complete evidence coverage under different source-turn budgets.}
Coverage and normalized AUC are reported as percentages.
AUC is computed over budgets of 10--120 source turns.
NEMORI top-10 denotes a hypothetical offline expansion of all retrieved episodes, not the original NEMORI answer context.
}
\label{tab:rq6_budget}
\setlength{\tabcolsep}{6pt}
\renewcommand{\arraystretch}{1.10}
\footnotesize

\begin{tabular*}{\textwidth}{@{\extracolsep{\fill}}lccccc@{}}
\toprule
\textbf{Retrieval Strategy}
& $\boldsymbol{B=20}$
& $\boldsymbol{B=40}$
& $\boldsymbol{B=80}$
& $\boldsymbol{B=120}$
& \textbf{AUC} \\
\midrule
\multicolumn{6}{l}{\textbf{(a) Claude Haiku 4.5}} \\
\midrule
NEMORI (actual)
& 9.3 & 15.0 & 16.2 & 16.2 & 14.5 \\
NEMORI top-10 (hypothetical)
& 12.3 & 32.3 & 57.5 & 65.6 & 44.4 \\
Matched Retrieval-Only
& 34.5 & 52.5 & 67.2 & 73.7 & 58.6 \\
\textbf{QEMem (Initial Retrieval)}
& \textbf{56.6} & \textbf{67.9}
& \textbf{80.4} & \textbf{82.6}
& \textbf{73.1} \\
\midrule
\multicolumn{6}{l}{\textbf{(b) Claude Opus 4.5}} \\
\midrule
NEMORI (actual)
& 9.3 & 15.0 & 16.2 & 16.2 & 14.5 \\
NEMORI top-10 (hypothetical)
& 12.3 & 32.3 & 57.5 & 65.6 & 44.4 \\
Matched Retrieval-Only
& 34.5 & 52.5 & 67.1 & 73.1 & 58.5 \\
\textbf{QEMem (Initial Retrieval)}
& \textbf{58.1} & \textbf{69.4}
& \textbf{81.0} & \textbf{83.0}
& \textbf{74.1} \\
\bottomrule
\end{tabular*}
\end{table*}

\begin{figure*}[t]
    \centering
    \includegraphics[width=\textwidth]{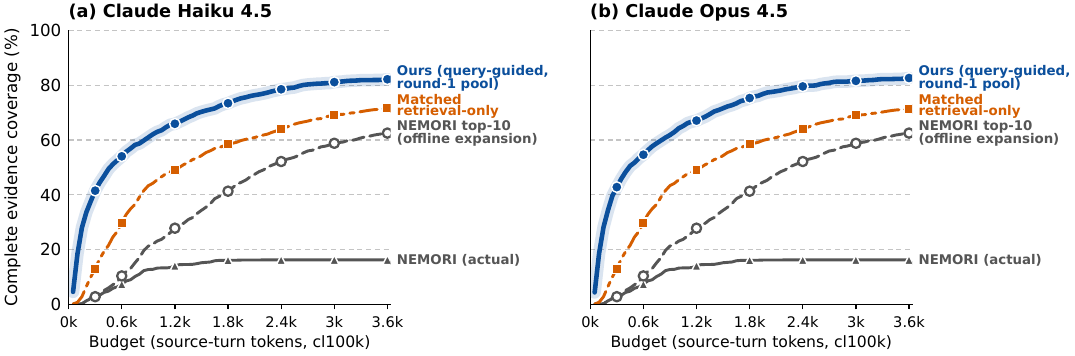}
    \caption{
    \textbf{Complete evidence recovery under source-turn token budgets.}
    \textbf{(a)} Claude Haiku 4.5 and
    \textbf{(b)} Claude Opus 4.5.
    Budgets are measured using the \texttt{cl100k\_base} tokenizer over original dialogue turns exposed verbatim.
    These counts exclude episodic summaries and semantic-memory text and are not equivalent to total answer-prompt tokens.
    The relative advantage of QEMem's initial retrieval pool remains consistent with the source-turn-budget analysis in Figure~\ref{fig:rq6_budget}.
    }
    \label{fig:rq6_token_budget}
\end{figure*}

\paragraph{Robustness across Token Budgets.}
To examine whether the results depend on source-turn counts, we repeat the analysis using the tokenized lengths of exposed original dialogue turns. Figure~\ref{fig:rq6_token_budget} shows complete evidence coverage under source-turn token budgets from 300 to 3,600 tokens. The results closely reproduce the turn-budget findings. QEMem improves normalized coverage AUC over matched retrieval-only by 16.1 and 17.1 percentage points under Haiku and Opus, respectively. We additionally evaluate an inclusive budget rule that allows the next retrieved episode to exceed the nominal budget, thereby reducing the disadvantage of episode-level retrieval granularity. Under this more permissive rule, QEMem still improves complete coverage by approximately 9--10 points at budgets of 20--80 turns. Similar results are obtained when restricting the analysis to questions without refinement, for which the initial retrieval pool is exactly recoverable from the saved traces. These checks indicate that the observed advantage is not solely an artifact of candidate granularity, budget truncation, or approximate retrieval-round reconstruction.

\paragraph{Query Planning versus Multi-Granular Retrieval.}
The budgeted analysis also helps distinguish the contribution of query planning from that of fine-grained evidence retrieval. Using the existing 400-question ablation subset, we compare the retrieval outputs produced with and without query planning while retaining the multi-granular retrieval mechanism. Under a tight budget of 20 source turns, removing query planning actually improves complete evidence coverage by 11.8 and 9.8 percentage points under Haiku and Opus, respectively. However, when the complete retrieval lists are considered, query planning yields higher coverage by 3.8 and 3.0 points, with larger improvements on multi-hop and cross-session questions. This suggests that direct query retrieval can prioritize immediately relevant evidence under tight budgets, whereas query decomposition becomes more beneficial when sufficient capacity is available to recover complementary information across multiple information needs. Since this ablation also removes planner-dependent budget adaptation and is evaluated on a smaller subset, the results do not isolate query planning as an independent causal factor. More broadly, the primary budget-efficiency advantage is most directly supported by QEMem's multi-granular source-turn retrieval rather than query planning alone.

\paragraph{Refinement and Source Reification.}
Beyond the initial retrieval pool, QEMem further improves evidence availability through selective refinement and source-grounded evidence construction. Under Haiku, complete source visibility increases from 82.6\% in the initial retrieval pool to 85.4\% after refinement, and reaches 91.6\% in the final answer context. The corresponding progression under Opus is 83.1\%, 84.3\%, and 91.7\%. The final increase arises from additional original turns brought into the answer context through ViewCard source references. This distinction is important: the budget curves measure ordered retrieval-prefix coverage, whereas the final answer context additionally incorporates evidence selected through source-grounded reification. Consequently, the final visibility values are reported as separate endpoints rather than points on the retrieval-budget curves.

\paragraph{Limitations.}
Several considerations qualify this analysis. First, truncating previously saved retrieval results does not reproduce the behavior of a system rerun under smaller retrieval budgets; query planning, evidence construction, and refinement decisions could change. Second, budgets count original source turns or their tokenized lengths, not the total prompt size, which additionally includes episodic summaries and semantic memories. Third, retrieval granularity differs across strategies: QEMem retrieves fine-grained source turns, whereas episode-based baselines expose turns in larger groups. Although the inclusive-budget sensitivity analysis mitigates this concern, it does not eliminate all differences in retrieval behavior. Finally, the retrieval variants differ in ranking mechanisms, and annotated supporting turns are an imperfect proxy for all information sufficient to answer a question. Therefore, these results establish stronger recovery of annotated evidence within stored retrieval prefixes, rather than superior downstream accuracy under arbitrary inference budgets.

\subsection{Answer-Level Groundedness and Faithfulness}
\label{app:answer_groundedness}

\paragraph{Evaluation Protocol.}
While RQ4 evaluates the provenance and fidelity of intermediate evidence representations, we further investigate whether these properties extend to the final generated answers. We conduct an answer-level grounding audit using 240 stratified LoCoMo questions and the existing predictions from NEMORI, SimpleMem, MemR$^3$, and QEMem under Claude Opus 4.5. Questions are sampled across reasoning categories, evidence complexity, and answer correctness, with sampling weights applied to estimate population-level results. For each question, we construct a shared verification context from the original conversations, including annotated supporting turns, neighboring dialogue context, and additional passages retrieved through BM25. All methods are evaluated against the same source evidence. An independent Opus 4.5 judge examines each answer without access to method identity, reference answers, or existing correctness labels. Factual claims are classified as supported, unsupported, contradicted, or unverifiable. Claims initially lacking evidence undergo an additional retrieval pass over the original conversation before final classification. We distinguish claims constituting the final answer from auxiliary explanatory claims and report abstentions separately.

\begin{figure*}[t]
    \centering
    \includegraphics[width=\textwidth]{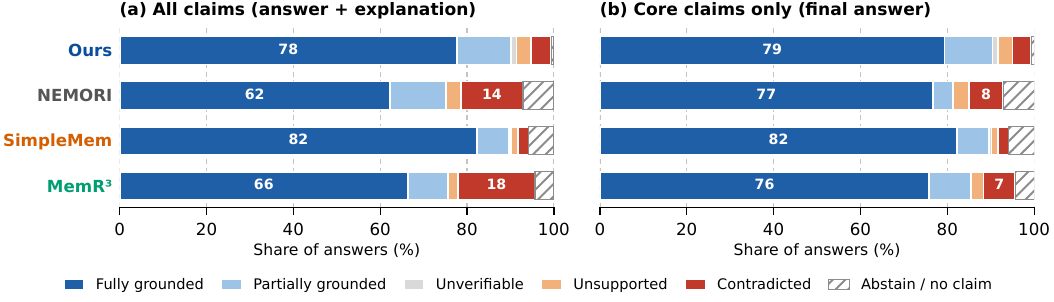}
    \caption{
    \textbf{Answer-level groundedness across memory systems on LoCoMo.}
    \textbf{(a)} Grounding classifications considering all factual claims, including auxiliary explanations.
    \textbf{(b)} Grounding classifications restricted to core claims constituting the final answer.
    Results are population-weighted estimates from 240 stratified questions using Claude Opus 4.5.
    The comparison distinguishes full grounding, partial grounding, unverifiable claims, unsupported claims, contradictions, and abstentions.
    Differences between the two panels highlight the influence of auxiliary explanatory content on answer-level faithfulness.
    }
    \label{fig:answer_groundedness}
\end{figure*}

\paragraph{Answer-Level Groundedness.}
Figure~\ref{fig:answer_groundedness} and Table~\ref{tab:answer_groundedness} summarize the results. QEMem achieves a fully grounded answer rate of 77.7\%, outperforming NEMORI (62.2\%) and MemR$^3$ (66.4\%) by 15.5 and 11.3 percentage points, respectively. Both improvements are statistically significant under paired bootstrap analysis, with 95\% confidence intervals of [7.2, 24.2] and [2.7, 20.1] percentage points. Moreover, the proportion of answers that are both correct and fully grounded increases to 73.2\%, compared with 55.4\% for NEMORI and 58.9\% for MemR$^3$. QEMem also produces substantially fewer answers containing source-contradicting assertions (4.6\%) than NEMORI (14.3\%) and MemR$^3$ (17.7\%), while maintaining an abstention rate of only 0.7\%. SimpleMem exhibits a slightly higher fully grounded rate (82.3\%), although its difference from QEMem is not statistically significant, and achieves a comparable correct-and-grounded rate (73.6\%). These findings suggest that QEMem maintains strong answer-level grounding alongside its accuracy improvements, particularly relative to methods that generate more extensive explanatory responses.

\begin{table*}[t]
\centering
\caption{
\textbf{Answer-level groundedness and faithfulness on LoCoMo.}
All metrics except claims per answer are percentages.
Results are population-weighted estimates from 240 stratified questions, evaluated using Claude Opus 4.5 predictions.
Fully grounded requires all checkable factual claims in an answer to be supported by source evidence.
Core-only grounding excludes auxiliary explanations. Grounding is evaluated over complete model outputs, including auxiliary reasoning and explanations; core-answer grounding considers only claims constituting the final answer.
}
\label{tab:answer_groundedness}
\setlength{\tabcolsep}{7pt}
\renewcommand{\arraystretch}{1.12}
\footnotesize

\begin{tabular*}{\textwidth}{@{\extracolsep{\fill}}lcccc@{}}
\toprule
\textbf{Metric}
& \textbf{NEMORI}
& \textbf{SimpleMem}
& \textbf{MemR$^3$}
& \textbf{QEMem} \\
\midrule
Fully Grounded $\uparrow$
& 62.2 & \textbf{82.3} & 66.4 & 77.7 \\
Fully or Partially Grounded $\uparrow$
& 75.1 & 89.7 & 75.5 & \textbf{90.2} \\
Correct \& Fully Grounded $\uparrow$
& 55.4 & \textbf{73.6} & 58.9 & 73.2 \\
Contradicted Answers $\downarrow$
& 14.3 & \textbf{2.4} & 17.7 & 4.6 \\
Abstention Rate $\downarrow$
& 7.1 & 5.8 & 4.3 & \textbf{0.7} \\
\midrule
Core-Claim Groundedness $\uparrow$
& 76.7 & \textbf{82.3} & 75.8 & 79.3 \\
Supported Claim Rate $\uparrow$
& 91.2 & \textbf{93.3} & 92.5 & 89.6 \\
Claims per Answer
& 4.90 & 2.38 & 5.75 & 2.73 \\
\bottomrule
\end{tabular*}
\end{table*}

\paragraph{The Role of Answer Verbosity.}
A closer examination reveals that the observed improvements in full-output groundedness are partly associated with differences in answer verbosity. NEMORI and MemR$^3$ generate an average of 4.90 and 5.75 factual claims per answer, respectively, compared with 2.73 for QEMem and 2.38 for SimpleMem. Their additional explanatory claims frequently introduce temporal inaccuracies, incorrect entity attributions, or unsupported details. Consequently, when evaluation is restricted to core claims constituting the final answer, the groundedness gap narrows substantially: QEMem achieves 79.3\%, compared with 76.7\% for NEMORI, 75.8\% for MemR$^3$, and 82.3\% for SimpleMem. These differences are not statistically significant. Similarly, claim-level support rates are broadly comparable across methods, ranging from 89.6\% to 93.3\%. Therefore, the evidence does not establish that individual QEMem claims are intrinsically more reliable. Instead, QEMem achieves stronger full-output grounding than the more verbose baselines by limiting unsupported auxiliary assertions while preserving high answer accuracy.

\paragraph{Correctness versus Groundedness.}
Answer correctness and evidence grounding capture distinct properties. An answer may be grounded in source material yet fail to fully answer the question, for example by omitting required items in a multi-entity query. Conversely, an answer may match the reference answer while introducing additional claims that are not supported by the original conversation. This distinction is reflected in the joint evaluation: QEMem obtains 73.2\% correct-and-fully-grounded answers, while 17.9\% of its answers are correct but not fully grounded under the strict all-claims criterion. Among correctly answered questions, its full-output groundedness is 80.4\%, compared with 70.1\% for NEMORI, 73.6\% for MemR$^3$, and 90.0\% for SimpleMem. These observations motivate evaluating both correctness and grounding rather than treating answer accuracy as a proxy for factual faithfulness.

\paragraph{Cross-Backbone Evidence.}
We additionally evaluate existing Haiku 4.5 answers on a smaller stratified subset of 68 questions. QEMem achieves a fully grounded rate of 74.4\%, compared with 54.0\% for NEMORI, 59.9\% for SimpleMem, and 51.4\% for MemR$^3$. Relative to NEMORI, the improvement is 20.4 percentage points (95\% CI: [8.4, 34.5]). The corresponding point estimates against SimpleMem and MemR$^3$ are also positive, but their question-level confidence intervals include zero. Furthermore, QEMem obtains a correct-and-fully-grounded rate of 74.0\%, compared with 44.2\%, 56.2\%, and 47.7\%, respectively. These results provide directional support for the findings under Opus 4.5, although the limited sample size prevents strong conclusions about cross-backbone consistency.

\paragraph{Reliability and Limitations.}
The grounding audit has several limitations. First, although all methods share the same original-dialogue verification context, the evidence packs may still omit relevant passages. Additional claim-specific retrieval mitigates this issue but cannot guarantee exhaustive source verification. Second, grounding judgments are produced by a single LLM judge and are sensitive to the interpretation of temporal relations, inferred facts, and question premises. In a stability check, 35 of 40 answer-level classifications remained unchanged after reordering the verification context. Manual inspection of 29 disputed cases identified nine erroneous judgments, indicating that the resulting estimates should be interpreted cautiously. Third, full-output groundedness depends on the number of factual claims: longer responses are more likely to contain at least one unsupported assertion. We therefore report core-only grounding, claim-level support, and answer length alongside the primary metrics. Finally, the main audit covers 240 stratified questions, while the Haiku analysis is limited to 68 questions. These results provide supplementary evidence that QEMem can produce well-grounded final answers, but do not establish a causal improvement in claim-level faithfulness attributable to evidence reification alone.

\subsection{Prompt Templates}
\label{app:prompt_templates}

We provide the prompt templates used by QEMem for query planning,
evidence construction, and answer generation. All prompts correspond
to the final QEMem-v2.3-clean implementation and are issued as a
single user message with temperature zero. Variables enclosed in
braces are populated dynamically at inference time. The illustrative
dates are fictional and do not overlap with the evaluation data.

\paragraph{Query-Guided Evidence Discovery.}
The query planner identifies the information requirements of
the question and generates complementary retrieval queries.
Its structured output also determines whether aggregation
or temporal reasoning is required.

\begin{QEMempromptbox}{Query Planning Prompt}
You are the query planner of a long-term conversational memory system.

The memory stores a long multi-session conversation between {speakers}. Sessions span {span}.

Question: {question}

Analyse what information is needed to answer the question and plan retrieval. Think about:
- which person(s), objects, events and attributes the question is about (resolve "he/she/they" to names when obvious),
- whether the answer needs several facts combined (multi-hop), facts aggregated over many sessions (lists, counts, "what activities/places/books..."), a date or time computation, or an inference about someone's preferences, personality or likely behaviour,
- which intermediate facts must be found first (e.g. to answer "What did X do after Y?" first find when Y happened).

Return ONLY a JSON object:
{
  "need": "<one sentence describing the information need>",
  "answer_type": "<one of: date, duration, count, list, yes_no, entity, description, reason, inference>",
  "aggregate": <true if the answer may be spread over multiple sessions, else false>,
  "temporal": <true if a date/time/order must be determined, else false>,
  "queries": ["<4-6 short, diverse search queries: sub-questions for intermediate facts, paraphrases with likely conversational wording, and key entity/keyword queries>"]
}
\end{QEMempromptbox}

The original question is combined with up to six planned queries
for retrieval across episodic memories, semantic memories, and
original conversation turns. The planner's information need is
subsequently provided to the evidence constructor.

\paragraph{Source-Grounded Evidence Reification.}
The evidence constructor transforms retrieved memories into
source-grounded ViewCards and jointly assesses whether the
current evidence is sufficient to answer the question. Importantly,
the sufficiency assessment and follow-up queries are generated
within the same evidence-construction call.

\begin{QEMempromptbox}{Evidence Construction Prompt}
You are the evidence constructor of a long-term conversational memory system.

Your job is NOT to answer the question but to build compact, source-grounded evidence cards that make the facts needed to answer it explicit.

Question: {question}
Information need: {need}

Candidate memories retrieved from the memory store (each has an ID; [E*] episode summaries, [S*] semantic facts, [T*] raw conversation turns with their session date):
{pool}
{previous}

Instructions:
1. Find every memory relevant to the question. For list/count/"what activities" questions collect ALL distinct instances across sessions; for multi-hop questions collect each intermediate fact.
2. For each relevant fact write a card. Prefer raw turns [T*] as sources because they are the original words; use [E*]/[S*] when no turn is available.
3. "quote" must be copied VERBATIM from the cited source (a short span, max ~30 words) - never paraphrase inside "quote".
4. "time": when the fact happened, resolved against the session date of the source: "yesterday" -> "11 March 2031" (absolute date); "last Friday"/"last weekend" -> "the Friday/weekend before 12 March 2031" (keep weekday-relative expressions relative to the session date, do not compute the calendar day); "last week" -> "the week before 12 March 2031"; "last year" -> "2030"; no relative expression -> the session date. Use "" if not time-relevant.
5. "relation": in a few words, how this card bears on the question (e.g. "answer candidate", "intermediate fact: date of Y", "supports inference: likes outdoors", "contradicts card 2").
6. Do not invent facts. If the memories conflict, make cards for both and note it in "relation".
7. Then judge whether the cards suffice. If key information is missing, give up to 3 follow-up search queries aimed precisely at the missing facts.

Return ONLY a JSON object:
{
  "cards": [{"claim": "<fact relevant to the question, self-contained, with names>", "sources": ["T3"], "quote": "<verbatim span>", "time": "<absolute time or empty>", "relation": "<role>"}],
  "sufficient": <true/false>,
  "missing": "<what is still missing, or empty>",
  "follow_up_queries": ["..."]
}
\end{QEMempromptbox}

The candidate pool is formatted with explicit memory identifiers
and source timestamps. During refinement, only newly retrieved
memories are supplied, together with a summary of previously
constructed evidence cards.

\begin{QEMempromptbox}{Candidate Memory Pool Format}
## Episodes
[E1] (<session date>) <episode title>: <episode content>
...

## Semantic facts
[S1] (learned in session of <session date>) <semantic fact>
...

## Conversation turns (grouped by session, chronological)
-- session <session date> --
[T1] <Speaker>: <original conversation turn>
[T2] <Speaker>: <original conversation turn>
...

-- session <session date> --
[T3] <Speaker>: <original conversation turn>
...
\end{QEMempromptbox}

When the constructor reports insufficient evidence, up to three
follow-up queries are used for another retrieval round. Previously
retrieved items are excluded, and the newly constructed cards are
merged into the existing evidence set. The additional text supplied
through the \texttt{\{previous\}} placeholder has the form:
\texttt{Evidence cards already constructed in the previous round}
followed by the existing claims and their source identifiers.

\paragraph{Evidence-Grounded Answer Generation.}
The answer model receives the constructed ViewCards, their
associated original source context, and the remaining retrieved
memories. Unverified cards are explicitly marked, while temporal
and aggregation instructions guide the final response.

\begin{QEMempromptbox}{Answer Generation Prompt}
You are an intelligent memory assistant answering questions about a long-term conversation between {speakers}, using evidence retrieved from memory.

# QUESTION
{question}

# EVIDENCE CARDS (constructed for this question; each claim is grounded in a verbatim quote from the cited source; cards marked UNVERIFIED could not be matched to their source and must be treated with caution)
{cards}

# CITED SOURCE CONTEXT (original conversation turns around the cited evidence, with session dates)
{sources}

# OTHER RETRIEVED MEMORIES (background; use them if the cards miss something)
{pool}

# INSTRUCTIONS
1. Base the answer on the evidence. Prefer verified cards and original turns; use episode summaries and semantic facts to fill gaps.

2. Time. Dates shown are session dates (the day the conversation took place). Express event times the way people resolve them against the session date:
   - "yesterday", "today", "this morning", "two days ago" -> the absolute calendar date (e.g. said on 12 March 2031 "yesterday" -> 11 March 2031);
   - "last Friday", "last Sunday", "on the weekend" (a weekday-relative expression) -> "the Friday before 12 March 2031" (relative to the session date; do NOT convert it into a calendar day yourself);
   - "last week", "last month", "a few days ago" -> "the week before 4 April 2031", "the month before ..." , "a few days before ...";
   - "last year", "next month" -> the year / month (e.g. said in 2031 "last year" -> 2030).
   If the event is reported without a relative expression, its time is the session date itself.
   When a card's time is of the form "the <weekday/week/weekend> before <date>", that is the resolved time: give it in exactly that form (episode summaries may contain mis-computed calendar days).

3. Lists / "what activities", "what books", "how many": aggregate ALL distinct items found across all sessions; count carefully and de-duplicate.

4. Questions about likely preferences, personality, opinions or hypotheticals ("Would X ...", "What might X ...", "Is X likely ..."): give the most plausible answer inferred from the evidence (start with "Probably yes"/"Probably no" for yes/no questions) rather than refusing. For questions about someone's traits or attributes, answer with a few short descriptive traits.

5. Always give the answer first. If the question's details (e.g. a date or name) do not exactly match the memories, answer with the memory that best matches the rest of the question - do not start with "there is no evidence".

6. Answer exactly what is asked; do not add extra facts, dates or explanations that are not asked for.

7. The final answer must be short and direct: a phrase, name, date, list or short sentence (ideally under 15 words).

First think briefly step by step (max ~120 words) under "REASONING:", then give the answer on the last line exactly as:

FINAL ANSWER: <answer>
\end{QEMempromptbox}

ViewCards are rendered with their claims, source identifiers,
quotations, temporal attributes, and evidential relations.
Cards that fail quotation verification are retained but marked
\texttt{UNVERIFIED}. The final answer is extracted from the
\texttt{FINAL ANSWER:} field.

\paragraph{Abstention Retry.}
If the initial answer matches the predefined abstention patterns,
QEMem performs at most one additional answering call using the same
evidence context, with the following instruction appended to the
original answer prompt.

\begin{QEMempromptbox}{Abstention Retry Prompt}
# NOTE
Your previous final answer was: "{prev}". A question usually refers to something that is in the conversation even when a detail in it (e.g. the year, month or a name) may be slightly wrong. Do not abstain: answer with the memory that best matches the question, as a short direct answer.
\end{QEMempromptbox}

\end{document}